%% file: main.tex
\pdfoutput=1
\documentclass[10pt]{article}

\usepackage[T1]{fontenc}
\usepackage[utf8]{inputenc}
\usepackage{lmodern}
\usepackage[margin=0.95in]{geometry}
\usepackage{microtype}

\usepackage{amsmath}      
\usepackage{booktabs}     
\usepackage{graphicx}     
\graphicspath{{../figures/}{figures/}{./}}

\usepackage[round,authoryear]{natbib}   
\usepackage{url}
\usepackage[hidelinks]{hyperref}

\hypersetup{
  pdftitle={Jev-as-a-Judge for RAG Claim Verification: Confidence-Gated Cascading
            Between a Non-Generative Verifier and a Frontier LLM},
  pdfauthor={Alden Do Rosario, Kirollos Atef},
  pdfsubject={cs.CL (primary); cs.AI (cross-list)},
  pdfkeywords={grounded claim verification, retrieval-augmented generation,
               LLM-as-a-judge, model cascades, selective prediction, calibration}
}

\IfFileExists{../generated/numbers.tex}
  {\input{../generated/numbers.tex}}
  {\input{numbers.tex}}

\newcommand{\paperinput}[1]{%
  \IfFileExists{sections/#1.tex}{\input{sections/#1.tex}}{\input{#1.tex}}}

\title{Jev-as-a-Judge for RAG Claim Verification:\\
Confidence-Gated Cascading Between a Non-Generative Verifier and a Frontier LLM}

\author{%
  Alden Do Rosario \qquad Kirollos Atef\\[2pt]
  CustomGPT.ai\\
  \texttt{alden@customgpt.ai} \qquad \texttt{kirollos@customgpt.ai}
}

\date{}

\begin{document}

\maketitle

\begin{abstract}
Retrieval-augmented generation asks one question at volume: does this evidence support this
claim? The answer is one bit, yet most pipelines obtain it from a text generator and pay
generation prices for the prose wrapped around it. We compare a non-generative decision
model, which returns a typed value from a declared enumeration together with a native
probability and no text, against a frontier LLM configured as an engineer would deploy it
for binary classification. Both answer the same \NPaired{} human-labelled claims from
LLM-AggreFact, so every comparison is within-item.

We report three things. First, the two systems are statistically indistinguishable on the
metric this task is conventionally scored by, a paired difference of
$\JevAstraDeltaSignedTwoDP$ points, Jev minus Astra, at a \CostRatioAstraOverJev-fold cost
difference. Second, that tie carries almost no information about deployment: the systems
disagree on \DisagreementPct\% of claims, and the cheaper one admits \JevFVR\% of the
claims annotators marked unsupported against the frontier model's \AstraFVR\%. A component
whose job is to catch unsupported claims should be judged on that rate, and the
conventional headline hides it. Third, because the errors differ they compose: routing the
claims the cheap model is confident about and escalating the rest reaches
\CascadeCrossfitBAccTwoDP{} at \$\CascadeCrossfitCost{} per thousand claims against
\AstraBAccTwoDP{} at \$\AstraCostPerK{}, with the threshold chosen out of fold. But a
matched random-escalation control at in-sample thresholds, spending the same budget by coin
flip, does as well, so what buys the saving is the escalation share and not the probability
that picks what to escalate. Measured against the cheap model it is built from, the cascade
costs \CostRatioCascadeOverJev-fold more for \CascadeCrossfitMinusJevPP{} points, 95\% CI
[$\CascadeJevDeltaCILo$, \CascadeJevDeltaCIHi], and it is the more permissive guard of the
two.

Equal accuracy conceals unequal errors. All predictions, prompts, pricing and code are
released.
\end{abstract}

\input{sections/setup.tex}

\input{sections/same-score.tex}

\input{sections/different-errors.tex}

\input{sections/cascade.tex}

\input{sections/close.tex}

{\small
\bibliographystyle{plainnat}
\bibliography{references}}

\end{document}

%% file: numbers.tex


\newcommand{\NPaired}{495}
\newcommand{\NDatasets}{11}
\newcommand{\NGoldSupported}{297}
\newcommand{\NGoldUnsupported}{198}
\newcommand{\GoldSupportedRatePct}{60.0}
\newcommand{\BootstrapResamples}{10000}


\newcommand{\SupportAggreFactCNNN}{45}
\newcommand{\SupportAggreFactCNNPos}{38}
\newcommand{\SupportAggreFactCNNNeg}{7}
\newcommand{\SupportAggreFactXSumN}{45}
\newcommand{\SupportAggreFactXSumPos}{18}
\newcommand{\SupportAggreFactXSumNeg}{27}
\newcommand{\SupportClaimVerifyN}{45}
\newcommand{\SupportClaimVerifyPos}{37}
\newcommand{\SupportClaimVerifyNeg}{8}
\newcommand{\SupportExpertQAN}{45}
\newcommand{\SupportExpertQAPos}{37}
\newcommand{\SupportExpertQANeg}{8}
\newcommand{\SupportFactCheckGPTN}{45}
\newcommand{\SupportFactCheckGPTPos}{7}
\newcommand{\SupportFactCheckGPTNeg}{38}
\newcommand{\SupportLfqaN}{45}
\newcommand{\SupportLfqaPos}{21}
\newcommand{\SupportLfqaNeg}{24}
\newcommand{\SupportRAGTruthN}{45}
\newcommand{\SupportRAGTruthPos}{41}
\newcommand{\SupportRAGTruthNeg}{4}
\newcommand{\SupportRevealN}{45}
\newcommand{\SupportRevealPos}{11}
\newcommand{\SupportRevealNeg}{34}
\newcommand{\SupportTofuEvalMediaSN}{45}
\newcommand{\SupportTofuEvalMediaSPos}{39}
\newcommand{\SupportTofuEvalMediaSNeg}{6}
\newcommand{\SupportTofuEvalMeetBN}{45}
\newcommand{\SupportTofuEvalMeetBPos}{38}
\newcommand{\SupportTofuEvalMeetBNeg}{7}
\newcommand{\SupportWiceN}{45}
\newcommand{\SupportWicePos}{10}
\newcommand{\SupportWiceNeg}{35}
\newcommand{\ThinnestMinorityClassN}{4}


\newcommand{\JevBAcc}{73.3}
\newcommand{\JevBAccTwoDP}{73.27}
\newcommand{\JevBAccCILo}{68.5}
\newcommand{\JevBAccCIHi}{77.9}

\newcommand{\JevRawAcc}{79.8}
\newcommand{\JevPooledBAcc}{79.3}
\newcommand{\JevFVR}{23.2}
\newcommand{\JevFVRTwoDP}{23.23}

\newcommand{\JevVerifiedPrecision}{84.1}
\newcommand{\JevMacroFOne}{79.1}
\newcommand{\JevMCC}{0.582}
\newcommand{\JevPredSupportedRate}{58.4}
\newcommand{\JevCostPerK}{0.04786}
\newcommand{\JevValidRatePct}{100.0}

\newcommand{\JevMeanInputTokens}{1,139}
\newcommand{\JevMeanOutputTokens}{35.4}
\newcommand{\JevMaxInputTokens}{12,977}

\newcommand{\JevMinOutputTokens}{35}
\newcommand{\JevMaxOutputTokens}{36}
\newcommand{\JevDistinctOutputTokenCounts}{2}
\newcommand{\JevBrier}{0.153}
\newcommand{\JevNLL}{1.133}
\newcommand{\JevNLLClipN}{10}
\newcommand{\JevNLLClipSharePct}{62}
\newcommand{\JevCalWorstP}{0.05}
\newcommand{\JevCalWorstObserved}{0.29}

\newcommand{\JevCalUnderConfidentAtOrBelow}{0.29}
\newcommand{\JevCalOverConfidentLo}{0.92}
\newcommand{\JevCalOverConfidentHi}{0.98}
\newcommand{\JevCalInteriorShareOfECEPct}{61}
\newcommand{\JevCalInteriorHi}{0.95}
\newcommand{\JevCalWorstConfidence}{0.95}
\newcommand{\JevCalBestP}{0.64}

\newcommand{\JevCalFirstAtOrBelowP}{0.64}
\newcommand{\JevCalBinsBelowDiagonalN}{6}

\newcommand{\JevCalZeroInLowestBinN}{50}
\newcommand{\JevNLLClipNatsEach}{34.5}
\newcommand{\JevNLLClipEpsMantissa}{1.0}
\newcommand{\JevNLLClipEpsExponent}{15}
\newcommand{\JevProbDecimals}{2}

\newcommand{\JevProbAtZeroN}{86}
\newcommand{\JevProbAtOneN}{26}
\newcommand{\JevProbStep}{0.01}
\newcommand{\JevProbHalfStep}{0.005}
\newcommand{\JevNLLConstantHalf}{0.693}
\newcommand{\JevBrierConstantHalf}{0.250}
\newcommand{\JevNLLAtHalfStepClip}{0.544}
\newcommand{\JevNLLAtEpsMicro}{0.715}
\newcommand{\JevEpsGridMicroMantissa}{1.0}
\newcommand{\JevEpsGridMicroExponent}{6}
\newcommand{\JevNLLAtEpsMilli}{0.575}
\newcommand{\JevEpsGridMilliMantissa}{1.0}
\newcommand{\JevEpsGridMilliExponent}{3}
\newcommand{\JevNLLCrossoverEpsMantissa}{2.9}
\newcommand{\JevNLLCrossoverEpsExponent}{6}
\newcommand{\JevECE}{0.099}
\newcommand{\JevECESensitivity}{0.107}
\newcommand{\JevAUROC}{0.868}
\newcommand{\JevAUPRC}{0.893}
\newcommand{\JevSelectiveAUROC}{0.722}
\newcommand{\JevSelectiveAUPRC}{0.892}
\newcommand{\JevBAccAggreFactCNN}{48.7}
\newcommand{\JevBAccAggreFactXSum}{67.6}
\newcommand{\JevBAccClaimVerify}{84.8}
\newcommand{\JevBAccExpertQA}{60.5}
\newcommand{\JevBAccFactCheckGPT}{64.8}
\newcommand{\JevBAccLfqa}{84.2}
\newcommand{\JevBAccRAGTruth}{92.7}
\newcommand{\JevBAccReveal}{72.9}
\newcommand{\JevBAccTofuEvalMediaS}{61.5}
\newcommand{\JevBAccTofuEvalMeetB}{80.5}
\newcommand{\JevBAccWice}{87.9}


\newcommand{\AstraBAccTwoDP}{73.83}
\newcommand{\AstraBAccCILo}{68.8}
\newcommand{\AstraBAccCIHi}{78.8}

\newcommand{\AstraRawAcc}{78.4}
\newcommand{\AstraPooledBAcc}{79.6}
\newcommand{\AstraFVR}{14.1}
\newcommand{\AstraFVRTwoDP}{14.14}
\newcommand{\AstraFVRCount}{28}
\newcommand{\AstraVerifiedPrecision}{88.6}
\newcommand{\AstraMacroFOne}{78.2}
\newcommand{\AstraMCC}{0.581}
\newcommand{\AstraPredSupportedRate}{49.7}
\newcommand{\AstraCostPerK}{8.93}
\newcommand{\AstraValidRatePct}{100.0}

\newcommand{\AstraMeanInputTokens}{843}
\newcommand{\AstraMeanOutputTokens}{10.0}
\newcommand{\AstraMaxInputTokens}{12,359}

\newcommand{\AstraMinOutputTokens}{4}
\newcommand{\AstraMaxOutputTokens}{134}
\newcommand{\AstraDistinctOutputTokenCounts}{47}
\newcommand{\AstraReasoningMedian}{0}

\newcommand{\AstraReasoningZeroN}{407}
\newcommand{\AstraBAccAggreFactCNN}{68.8}
\newcommand{\AstraBAccAggreFactXSum}{63.0}
\newcommand{\AstraBAccClaimVerify}{85.6}
\newcommand{\AstraBAccExpertQA}{46.1}
\newcommand{\AstraBAccFactCheckGPT}{67.5}
\newcommand{\AstraBAccLfqa}{81.8}
\newcommand{\AstraBAccRAGTruth}{87.8}
\newcommand{\AstraBAccReveal}{80.5}
\newcommand{\AstraBAccTofuEvalMediaS}{71.2}
\newcommand{\AstraBAccTofuEvalMeetB}{79.1}
\newcommand{\AstraBAccWice}{80.7}


\newcommand{\AstraHighBAccTwoDP}{74.11}
\newcommand{\AstraHighBAccCILo}{69.1}
\newcommand{\AstraHighBAccCIHi}{79.0}

\newcommand{\AstraHighFVRTwoDP}{13.13}
\newcommand{\AstraHighFVRCount}{26}

\newcommand{\AstraHighCostPerK}{10.70}

\newcommand{\AstraHighMeanInputTokens}{843}
\newcommand{\AstraHighMeanOutputTokens}{45.3}

\newcommand{\AstraHighMinOutputTokens}{4}
\newcommand{\AstraHighMaxOutputTokens}{519}
\newcommand{\AstraHighDistinctOutputTokenCounts}{109}
\newcommand{\AstraHighReasoningMedian}{21}

\newcommand{\AstraHighReasoningZeroN}{77}
\newcommand{\AstraHighBAccAggreFactCNN}{68.8}
\newcommand{\AstraHighBAccAggreFactXSum}{65.7}
\newcommand{\AstraHighBAccClaimVerify}{85.6}
\newcommand{\AstraHighBAccExpertQA}{46.1}
\newcommand{\AstraHighBAccFactCheckGPT}{67.5}
\newcommand{\AstraHighBAccLfqa}{86.0}
\newcommand{\AstraHighBAccRAGTruth}{87.8}
\newcommand{\AstraHighBAccReveal}{80.5}
\newcommand{\AstraHighBAccTofuEvalMediaS}{68.6}
\newcommand{\AstraHighBAccTofuEvalMeetB}{77.8}
\newcommand{\AstraHighBAccWice}{80.7}


\newcommand{\MiniCheckBAcc}{61.6}

\newcommand{\MiniCheckBAccCILo}{56.3}
\newcommand{\MiniCheckBAccCIHi}{66.7}
\newcommand{\MiniCheckBAccCIHalfWidthPP}{5.2}

\newcommand{\MiniCheckPublishedTestBAcc}{75.0}

\newcommand{\MiniCheckShortfallVsPublishedPP}{13.4}

\newcommand{\MiniCheckRetuningHeadroomPP}{0.9}
\newcommand{\MiniCheckGapToJevAtBestCutPP}{10.8}
\newcommand{\MiniCheckCutpointsSwept}{454}


\newcommand{\JevAstraDeltaPP}{0.6}
\newcommand{\JevAstraDeltaSignedPP}{-0.6}
\newcommand{\JevAstraDeltaSignedTwoDP}{-0.56}
\newcommand{\JevAstraDeltaCILo}{-5.6}
\newcommand{\JevAstraDeltaCIHi}{4.5}
\newcommand{\JevAstraMcNemarP}{0.49}
\newcommand{\JevAstraMcNemarStat}{34}

\newcommand{\JevAstraHighDeltaSignedPP}{-0.8}

\newcommand{\JevAstraHighDeltaCILo}{-5.9}
\newcommand{\JevAstraHighDeltaCIHi}{4.1}
\newcommand{\JevAstraHighMcNemarP}{0.48}
\newcommand{\JevAstraHighMcNemarStat}{33}
\newcommand{\AstraAstraHighDeltaPP}{0.3}
\newcommand{\AstraAstraHighDeltaSignedPP}{-0.3}

\newcommand{\AstraAstraHighDeltaCILo}{-1.3}
\newcommand{\AstraAstraHighDeltaCIHi}{0.7}
\newcommand{\AstraAstraHighMcNemarP}{1.00}
\newcommand{\AstraAstraHighMcNemarStat}{6}


\newcommand{\AgreementPct}{84.8}
\newcommand{\DisagreementPct}{15.2}
\newcommand{\JevOnlyCorrect}{41}
\newcommand{\AstraOnlyCorrect}{34}
\newcommand{\BothCorrect}{354}
\newcommand{\NeitherCorrect}{66}
\newcommand{\ErrorKappa}{0.54}

\newcommand{\NeitherExpectedIndependent}{21.6}
\newcommand{\JointFailureExcess}{3.1}
\newcommand{\DatasetsJevHigher}{6}
\newcommand{\DatasetsAstraHigher}{5}
\newcommand{\FVRGapPP}{9.1}

\newcommand{\FVRGapCILo}{4.55}
\newcommand{\FVRGapCIHi}{13.64}
\newcommand{\FVRGapMcNemarP}{0.000534}
\newcommand{\FVRJevOnlyFalseVerifies}{22}
\newcommand{\FVRAstraOnlyFalseVerifies}{4}
\newcommand{\FVRBothFalseVerify}{24}

\newcommand{\WidestTwoSumPP}{-5.75}
\newcommand{\RemainingNineSumPP}{-0.36}
\newcommand{\NRemainingDatasets}{9}
\newcommand{\WidestTwoShareOfSumPct}{94}
\newcommand{\SumAbsDeltasPP}{75.5}
\newcommand{\SignedSumDeltasPP}{-6.12}
\newcommand{\WidestTwoSumAbsPP}{34.5}
\newcommand{\RemainingSumAbsPP}{41.1}
\newcommand{\MaxAbsRemainingPP}{9.6}
\newcommand{\MeanAbsRemainingPP}{4.6}
\newcommand{\WidestTwoCancellationPct}{83}
\newcommand{\RemainingCancellationPct}{99}
\newcommand{\DeltaAggreFactCNNPP}{-20.11}
\newcommand{\DeltaExpertQAPP}{14.36}


\newcommand{\JevRerunFlips}{7}

\newcommand{\JevRerunBAccTwoDP}{72.44}
\newcommand{\JevRerunDeltaPP}{0.8}

\newcommand{\JevRerunFalseVerifiedRAGTruth}{1}

\newcommand{\JevRerunBAccRAGTruth}{79.0}


\newcommand{\PromptParityFlips}{9}
\newcommand{\PromptParityPreBAcc}{73.06}
\newcommand{\PromptParityPostBAcc}{73.83}
\newcommand{\PromptParityPreFVR}{14.1}
\newcommand{\PromptParityPostFVR}{14.1}
\newcommand{\PromptParityFVRGapShiftPP}{0.00}
\newcommand{\PromptParityFlipsOnUnsupported}{0}
\newcommand{\PromptParityFlipsOnSupported}{9}
\newcommand{\PromptParityFlipsToSupported}{7}
\newcommand{\PromptParityFlipsToUnsupported}{2}
\newcommand{\PromptParityHighFlips}{5}
\newcommand{\PromptParityHighPreBAcc}{73.93}
\newcommand{\PromptParityHighPostBAcc}{74.11}
\newcommand{\PromptParityHighPreFVR}{12.6}
\newcommand{\PromptParityHighPostFVR}{13.1}
\newcommand{\PromptParityHighFlipsOnUnsupported}{1}

\newcommand{\CascadeAstraHighFVRPreParityDeltaPP}{6.60}
\newcommand{\CascadeAstraHighFVRPreParityMcNemarPMedian}{0.00024}
\newcommand{\CascadeAstraHighFVRPreParityRepeatsPBelowFive}{200}


\newcommand{\SamePromptRepeatCompared}{411}
\newcommand{\SamePromptRepeatExcluded}{84}
\newcommand{\SamePromptRepeatFlips}{0}

\newcommand{\SamePromptRepeatBAcc}{77.55}
\newcommand{\SamePromptRepeatBAccOther}{77.55}
\newcommand{\SamePromptRepeatDeltaPP}{0.00}
\newcommand{\AstraTokenCap}{256}
\newcommand{\AstraHighTokenCap}{4096}
\newcommand{\SamePromptRepeatCappedTokenCap}{16}
\newcommand{\SamePromptRepeatTokenCap}{256}
\newcommand{\SamePromptRepeatComparedReasoningZero}{395}

\newcommand{\PromptParityFlipsInCappedNoAnswer}{9}


\newcommand{\CostRatioAstraOverJev}{187}
\newcommand{\CostRatioAstraHighOverJev}{224}
\newcommand{\CostRatioAstraOverCascade}{3.06}


\newcommand{\JevInputPricePerMtok}{0.042}
\newcommand{\AstraInputPricePerMtok}{10.00}
\newcommand{\AstraOutputPricePerMtok}{50.00}
\newcommand{\InputPriceRatio}{238.1}
\newcommand{\InputTokenRatioJevOverAstra}{1.351}
\newcommand{\AstraOutputUplift}{1.059}
\newcommand{\RatioFromPriceAndTokens}{176.2}
\newcommand{\JevCostPerKIfOutputBilled}{0.04934}
\newcommand{\CostRatioIfOutputBilled}{181}
\newcommand{\FreeOutputShareOfRatioPct}{3}


\newcommand{\PublishedBespokeMiniCheckTestBAcc}{77.4}
\newcommand{\PublishedClaudeThreeFiveSonnetTestBAcc}{77.2}
\newcommand{\PublishedMiniCheckFlanTFiveLTestBAcc}{75.0}
\newcommand{\PublishedGptFourOTestBAcc}{75.9}
\newcommand{\PublishedLeaderboardRows}{4}


\newcommand{\GateAtFiftyCoverage}{100.0}
\newcommand{\GateAtFiftyBAcc}{73.3}
\newcommand{\GateAtFiftyVerifiedPrecision}{84.1}
\newcommand{\GateAtFiftyFVR}{23.2}
\newcommand{\GateAtFiftyNDatasets}{11}

\newcommand{\GateAtSeventyCoverage}{89.1}
\newcommand{\GateAtSeventyBAcc}{75.5}
\newcommand{\GateAtSeventyVerifiedPrecision}{86.0}
\newcommand{\GateAtSeventyFVR}{20.9}
\newcommand{\GateAtSeventyNDatasets}{11}

\newcommand{\GateAtEightyCoverage}{82.8}
\newcommand{\GateAtEightyBAcc}{77.2}
\newcommand{\GateAtEightyVerifiedPrecision}{87.4}
\newcommand{\GateAtEightyFVR}{18.7}
\newcommand{\GateAtEightyNDatasets}{11}

\newcommand{\GateAtNinetyCoverage}{69.1}
\newcommand{\GateAtNinetyBAcc}{75.8}
\newcommand{\GateAtNinetyVerifiedPrecision}{88.1}
\newcommand{\GateAtNinetyFVR}{16.8}
\newcommand{\GateAtNinetyNDatasets}{11}

\newcommand{\GateAtNinetyFiveCoverage}{55.8}
\newcommand{\GateAtNinetyFiveBAcc}{80.4}
\newcommand{\GateAtNinetyFiveVerifiedPrecision}{91.3}
\newcommand{\GateAtNinetyFiveFVR}{10.7}
\newcommand{\GateAtNinetyFiveNDatasets}{11}

\newcommand{\GateAtNinetySevenCoverage}{48.7}
\newcommand{\GateAtNinetySevenBAcc}{81.0}
\newcommand{\GateAtNinetySevenVerifiedPrecision}{93.5}
\newcommand{\GateAtNinetySevenFVR}{7.2}
\newcommand{\GateAtNinetySevenNDatasets}{11}

\newcommand{\GateAtNinetyNineCoverage}{35.2}
\newcommand{\GateAtNinetyNineBAcc}{87.9}
\newcommand{\GateAtNinetyNineVerifiedPrecision}{97.1}
\newcommand{\GateAtNinetyNineFVR}{2.1}
\newcommand{\GateAtNinetyNineNDatasets}{9}
\newcommand{\GateAtNinetyNineNDatasetsDropped}{2}


\newcommand{\CascadeAtSeventyBAccTwoDP}{74.81}

\newcommand{\CascadeAtSeventyCost}{1.00}
\newcommand{\CascadeAtSeventyFVR}{21.2}

\newcommand{\CascadeAtEightyBAccTwoDP}{75.82}
\newcommand{\CascadeAtEightyEscalated}{17.2}
\newcommand{\CascadeAtEightyCost}{1.62}
\newcommand{\CascadeAtEightyFVR}{20.2}

\newcommand{\CascadeAtNinetyBAccTwoDP}{74.01}

\newcommand{\CascadeAtNinetyCost}{2.94}
\newcommand{\CascadeAtNinetyFVR}{19.2}

\newcommand{\CascadeAtNinetyFiveBAccTwoDP}{74.49}

\newcommand{\CascadeAtNinetyFiveCost}{4.56}
\newcommand{\CascadeAtNinetyFiveFVR}{16.7}

\newcommand{\CascadeAtNinetySevenBAccTwoDP}{74.82}

\newcommand{\CascadeAtNinetySevenCost}{5.21}
\newcommand{\CascadeAtNinetySevenFVR}{15.2}

\newcommand{\CascadeAtNinetyNineBAcc}{75.3}
\newcommand{\CascadeAtNinetyNineBAccTwoDP}{75.30}
\newcommand{\CascadeAtNinetyNineEscalated}{64.8}
\newcommand{\CascadeAtNinetyNineCost}{6.43}
\newcommand{\CascadeAtNinetyNineFVR}{13.6}
\newcommand{\CascadeAtNinetyNineFVRCount}{27}

\newcommand{\CascadeAtNinetyNineFVRVsAstraMcNemarP}{1.00}
\newcommand{\CascadeAtNinetyNineMinusAstraPP}{1.5}

\newcommand{\CascadeInSampleBAccTwoDP}{75.82}
\newcommand{\CascadeInSampleEscalated}{17.2}
\newcommand{\CascadeInSampleCost}{1.62}
\newcommand{\CascadeInSampleThreshold}{0.80}


\newcommand{\CascadeCrossfitCILo}{72.1}
\newcommand{\CascadeCrossfitCIHi}{75.8}
\newcommand{\CascadeCrossfitEscalated}{29.9}
\newcommand{\CascadeCrossfitCost}{2.92}

\newcommand{\CascadeCrossfitFVRTwoDP}{19.22}

\newcommand{\CascadeAstraDeltaPP}{0.27}
\newcommand{\CascadeAstraDeltaCILo}{-3.85}
\newcommand{\CascadeAstraDeltaCIHi}{4.15}
\newcommand{\CascadeAstraDeltaCIWidthPP}{8.00}
\newcommand{\CascadeAstraMcNemarPMedian}{0.20}
\newcommand{\CascadeAstraMcNemarPMin}{0.01}
\newcommand{\CascadeAstraMcNemarPMax}{1.00}
\newcommand{\CascadeAstraDiscordantMedian}{34}
\newcommand{\CascadeAstraRepeatsPBelowFive}{16}

\newcommand{\CascadeAstraRepeatsPBelowFiveForCascade}{16}
\newcommand{\CascadeAstraMcNemarFavoursCascade}{198}
\newcommand{\CascadeAstraMcNemarFavoursAstra}{0}
\newcommand{\CascadeAstraRepeatsMacroBelowAstra}{68}
\newcommand{\CascadeAstraBootstrapDrawsNegativePct}{43.1}


\newcommand{\CascadeAstraHighDeltaPP}{-0.01}
\newcommand{\CascadeAstraHighDeltaCILo}{-4.15}
\newcommand{\CascadeAstraHighDeltaCIHi}{3.90}
\newcommand{\CascadeAstraHighMcNemarPMedian}{0.23}
\newcommand{\CascadeAstraHighMcNemarPMin}{0.02}
\newcommand{\CascadeAstraHighMcNemarPMax}{1.00}


\newcommand{\CostRatioCascadeOverJev}{61}
\newcommand{\CascadeCrossfitMinusJevPP}{0.83}

\newcommand{\CascadeJevDeltaPP}{0.83}
\newcommand{\CascadeJevDeltaCILo}{-3.36}
\newcommand{\CascadeJevDeltaCIHi}{4.86}
\newcommand{\CascadeJevMcNemarPMedian}{0.86}

\newcommand{\CascadeJevRepeatsMacroBelowJev}{29}

\newcommand{\SelectionOptimismPP}{1.72}
\newcommand{\CrossfitRepeats}{200}
\newcommand{\CrossfitFolds}{5}
\newcommand{\CrossfitThresholdFits}{1000}


\newcommand{\MultiplicityFamilyN}{13}
\newcommand{\NClearingUncorrected}{3}
\newcommand{\MultiplicitySurvivingN}{2}
\newcommand{\MultiplicityMedianMembersN}{6}
\newcommand{\FVRGapHolmBreakFamilyN}{94}
\newcommand{\CascadeGuardHolmClearsAtFamilyN}{11}
\newcommand{\CascadeGuardHolmPAtThatFamily}{0.044}
\newcommand{\CascadeGuardHolmPExcludingMiniCheck}{0.057}
\newcommand{\CascadeGuardHolmMembersAboveN}{8}
\newcommand{\CascadeGuardHolmBreakDroppedN}{2}
\newcommand{\AstraAstraHighMcNemarPHolm}{1.000}

\newcommand{\CascadeAstraMcNemarPHolm}{1.000}
\newcommand{\CascadeAstraFVRMcNemarPHolm}{0.057}
\newcommand{\CascadeAstraHighMcNemarPHolm}{1.000}
\newcommand{\CascadeAstraHighFVRMcNemarPHolm}{0.005}
\newcommand{\CascadeAtNinetyNineFVRVsAstraMcNemarPHolm}{1.000}
\newcommand{\CascadeJevMcNemarPHolm}{1.000}
\newcommand{\CascadeJevFVRMcNemarPHolm}{0.459}
\newcommand{\FVRGapMcNemarPHolm}{0.005}
\newcommand{\JevAstraMcNemarPHolm}{1.000}
\newcommand{\JevAstraHighMcNemarPHolm}{1.000}

\newcommand{\JevMiniCheckMcNemarPMantissa}{1.6}
\newcommand{\JevMiniCheckMcNemarPExponent}{16}
\newcommand{\JevMiniCheckMcNemarPHolmMantissa}{2.1}
\newcommand{\JevMiniCheckMcNemarPHolmExponent}{15}
\newcommand{\AstraMiniCheckMcNemarPMantissa}{4.0}
\newcommand{\AstraMiniCheckMcNemarPExponent}{14}
\newcommand{\AstraMiniCheckMcNemarPHolmMantissa}{4.8}
\newcommand{\AstraMiniCheckMcNemarPHolmExponent}{13}


\newcommand{\CascadeAstraFVRDeltaPP}{5.08}
\newcommand{\CascadeAstraFVRDeltaCILo}{1.52}
\newcommand{\CascadeAstraFVRDeltaCIHi}{9.09}
\newcommand{\CascadeAstraFVRMcNemarPMedian}{0.00635}

\newcommand{\CascadeAstraFVRRepeatsPBelowFive}{193}

\newcommand{\CascadeJevFVRDeltaPP}{-4.01}
\newcommand{\CascadeJevFVRDeltaCILo}{-8.08}
\newcommand{\CascadeJevFVRDeltaCIHi}{-0.51}
\newcommand{\CascadeJevFVRMcNemarPMedian}{0.05737}

\newcommand{\CascadeJevFVRRepeatsPBelowFive}{76}
\newcommand{\CascadeJevFVRRepeatsFavouringCascade}{200}
\newcommand{\CascadeAstraHighFVRDeltaPP}{6.09}
\newcommand{\CascadeAstraHighFVRDeltaCILo}{2.53}
\newcommand{\CascadeAstraHighFVRDeltaCIHi}{10.10}
\newcommand{\CascadeAstraHighFVRMcNemarPMedian}{0.00049}

\newcommand{\CascadeAstraHighFVRRepeatsPBelowFive}{200}


\newcommand{\CrossfitSEMBAccPP}{0.062}
\newcommand{\CrossfitSEMCost}{0.048}
\newcommand{\CrossfitSEMEscalatedPP}{0.467}
\newcommand{\CrossfitSEMFVRPP}{0.068}
\newcommand{\RepeatsReproducingInSample}{13}
\newcommand{\RepeatsReproducingInSamplePct}{6.5}

\newcommand{\ThresholdPickedFifty}{1}
\newcommand{\ThresholdPickedSeventy}{47}
\newcommand{\ThresholdPickedEighty}{669}
\newcommand{\ThresholdPickedNinety}{0}
\newcommand{\ThresholdPickedNinetyFive}{4}
\newcommand{\ThresholdPickedNinetySeven}{26}
\newcommand{\ThresholdPickedNinetyNine}{253}


\newcommand{\CascadeCrossfitBAccTwoDP}{74.10}
\newcommand{\OracleBAccTwoDP}{82.91}


\newcommand{\RandomAtSeventyEscalated}{10.9}

\newcommand{\RandomAtSeventyBAccTwoDP}{73.34}

\newcommand{\RandomAtSeventyCILoTwoDP}{71.95}
\newcommand{\RandomAtSeventyCIHiTwoDP}{74.94}

\newcommand{\RandomAtEightyEscalated}{17.2}

\newcommand{\RandomAtEightyBAccTwoDP}{73.31}

\newcommand{\RandomAtEightyCILoTwoDP}{71.39}
\newcommand{\RandomAtEightyCIHiTwoDP}{75.30}

\newcommand{\RandomAtNinetyEscalated}{30.9}

\newcommand{\RandomAtNinetyBAccTwoDP}{73.36}

\newcommand{\RandomAtNinetyCILoTwoDP}{71.13}
\newcommand{\RandomAtNinetyCIHiTwoDP}{76.01}
\newcommand{\RandomAtNinetyCost}{2.81}

\newcommand{\RandomAtNinetyFiveEscalated}{44.2}

\newcommand{\RandomAtNinetyFiveBAccTwoDP}{73.43}

\newcommand{\RandomAtNinetyFiveCILoTwoDP}{70.76}
\newcommand{\RandomAtNinetyFiveCIHiTwoDP}{76.03}

\newcommand{\RandomAtNinetySevenEscalated}{51.3}

\newcommand{\RandomAtNinetySevenBAccTwoDP}{73.51}

\newcommand{\RandomAtNinetySevenCILoTwoDP}{71.40}
\newcommand{\RandomAtNinetySevenCIHiTwoDP}{76.03}

\newcommand{\RandomAtNinetyNineEscalated}{64.8}

\newcommand{\RandomAtNinetyNineBAccTwoDP}{73.59}

\newcommand{\RandomAtNinetyNineCILoTwoDP}{71.27}
\newcommand{\RandomAtNinetyNineCIHiTwoDP}{75.82}

\newcommand{\NRandomBudgetsTested}{6}
\newcommand{\RandomVsAstraDeltaPP}{-0.47}
\newcommand{\RandomVsAstraDeltaCILo}{-5.44}
\newcommand{\RandomVsAstraDeltaCIHi}{4.41}
\newcommand{\RandomVsAstraMcNemarPMedian}{0.63}
\newcommand{\RandomVsAstraRepeatsPBelowFive}{1}
\newcommand{\NRandomBudgetsConfidenceWins}{6}
\newcommand{\ConfMinusRandomMinPP}{0.65}
\newcommand{\ConfMinusRandomMaxPP}{2.51}
\newcommand{\OracleBAcc}{82.9}
\newcommand{\OracleEscalated}{20.2}
\newcommand{\OracleCost}{1.60}
\newcommand{\OracleHeadroomPP}{9.6}

\newcommand{\OracleFVRTwoDP}{12.12}

%% file: setup.tex

\section{Introduction}
\label{sec:intro}

A retrieval-augmented generation pipeline asks one question over and over, at whatever volume
its traffic dictates: does this evidence support this claim? The answer is one bit. Most
systems produce it with a text generator that writes a short verdict for downstream code to
parse back into a boolean, billed at generation prices. The bit is what the pipeline wanted;
the essay around it is overhead paid on every claim, forever.

Another kind of machine exists. A non-generative decision model takes the claim and the
evidence as typed state, returns one value from a declared enumeration with a probability over
it, and emits no prose. We evaluate \texttt{jev-1.13.0}, a commercial model of this kind,
against \texttt{gpt-6-astra}, a frontier LLM configured as a production engineer would deploy
it for binary classification, on \NPaired{} human-labelled claims from the
\NDatasets{} component datasets of LLM-AggreFact \citep{tang2024minicheck}. Every claim goes to both systems, so every comparison here is
within-item, at a cost difference between the arms of \CostRatioAstraOverJev-fold.

Their headline numbers land within a point of each other (Table~\ref{tab:headline}): a
paired difference of $\JevAstraDeltaSignedTwoDP$ points on Jev minus Astra, 95\% CI
[$\JevAstraDeltaCILo$, $+\JevAstraDeltaCIHi$], exact McNemar $p = \JevAstraMcNemarP$. That
tie is the least interesting fact we found, because the two systems do not make the same
mistakes. They disagree on \DisagreementPct\% of individual claims, and where the gold
label says a claim is unsupported the decision model waves it through \JevFVRTwoDP\% of the
time against the frontier model's \AstraFVRTwoDP\%, a gap of \FVRGapPP{} points. One of
these is a materially more permissive guard, and the headline metric cannot tell you which.

Errors that differ can be composed. Route each claim to the cheap model, accept its verdict
when its native probability is confident, and escalate the rest. With the threshold chosen
out of sample by cross-fitting, that cascade reaches \CascadeCrossfitBAccTwoDP\% at
\$\CascadeCrossfitCost{} per thousand claims against \AstraBAccTwoDP\% at \$\AstraCostPerK{}
for the frontier model alone. Escalating the same share at random costs
\$\RandomAtNinetyCost{} and is no further from the frontier arm, so what pays is the share
escalated rather than the signal that picks it (Section~\ref{subsec:random-control}).
Choosing the threshold on the data it is scored on, the identical procedure reports
\CascadeInSampleBAccTwoDP\%; the \SelectionOptimismPP{} points between them are selection,
which tuned cascade results are routinely published without. What it does not buy is a
tighter guard, since on the claims it keeps it is the cheap model: it false-verifies
\CascadeCrossfitFVRTwoDP\% against \AstraFVRTwoDP\%. Table~\ref{tab:headline} prints the
column on which this cost result loses.

\begin{table}[t]
\centering
\small
\begin{tabular}{lccc}
\toprule
Configuration & Macro BAcc [interval] & False verification & Cost / 1k claims \\
\midrule
\texttt{jev-1.13.0}                  & \JevBAccTwoDP\ [\JevBAccCILo, \JevBAccCIHi]                   & \JevFVRTwoDP\%       & \$\JevCostPerK \\
\texttt{gpt-6-astra}, effort low     & \AstraBAccTwoDP\ [\AstraBAccCILo, \AstraBAccCIHi]             & \AstraFVRTwoDP\%     & \$\AstraCostPerK \\
\texttt{gpt-6-astra}, effort high    & \AstraHighBAccTwoDP\ [\AstraHighBAccCILo, \AstraHighBAccCIHi] & \AstraHighFVRTwoDP\% & \$\AstraHighCostPerK \\
\midrule
Cascade, cross-fitted                & \CascadeCrossfitBAccTwoDP\ [\CascadeCrossfitCILo, \CascadeCrossfitCIHi]$^{\dagger}$ & \CascadeCrossfitFVRTwoDP\%$^{\ddagger}$ & \$\CascadeCrossfitCost \\
\bottomrule
\end{tabular}
\caption{Two systems at three configurations, plus the cross-fitted cascade, on the same
\NPaired{} claims. The bracket column holds two kinds of interval, which is why its heading
does not say confidence interval: on the three single-system rows it is a 95\% paired
stratified bootstrap confidence interval over claims, and on the cascade row it is not. The
point estimates and the false-verification rates are printed to two decimals, because the
two footnotes below print differences of them and a reader is entitled to reproduce those
differences, and because the high-effort frontier row and the cascade row are different
quantities that round to the same value at one decimal. $^{\dagger}$The cascade interval is the spread over cross-fitting repeats and
carries threshold-selection noise rather than sampling uncertainty. The sampling comparison
against the frontier arm is made separately in Section~\ref{sec:cascade-inference}: the
difference is \CascadeAstraDeltaPP\ points, 95\% CI [${\CascadeAstraDeltaCILo}$,
\CascadeAstraDeltaCIHi], so this sample does not resolve which of the two is more accurate.
$^{\ddagger}$The cascade's false-verification rate is cross-fitted in the same loop as its
accuracy. It sits \CascadeAstraFVRDeltaPP\ points above the frontier arm's, 95\% CI
[\CascadeAstraFVRDeltaCILo, \CascadeAstraFVRDeltaCIHi] on a paired bootstrap over the
gold-unsupported claims, with a median exact McNemar $p$ of
\CascadeAstraFVRMcNemarPMedian, and $\CascadeAstraFVRMcNemarPHolm$ adjusted over the
\MultiplicityFamilyN-test family. Unlike the accuracy comparison in the dagger note, this one
meets the separation criterion of Section~\ref{sec:setup} uncorrected, and sits on the $0.05$
boundary once corrected, so the cascade is the cheaper configuration and the more permissive
guard at once. Costs are uncached
standard-tier prices applied to measured token counts.}
\label{tab:headline}
\end{table}

Our contributions are four.

\begin{enumerate}
\item A paired evaluation of a non-generative decision model against a frontier LLM on
grounded claim verification, on which the two are statistically indistinguishable at a
\CostRatioAstraOverJev-fold difference in cost (Section~\ref{sec:same-score}).

\item An error-level characterization of that tie: the decision model is right where the
frontier model is wrong \JevOnlyCorrect{} times against \AstraOnlyCorrect{} the other way,
and their false-verification rates differ by \FVRGapPP{} points
(Section~\ref{sec:different-errors}).

\item A probability-provenance taxonomy (native, logprob-derived, none) and the distinction
that governs routing: how well a probability discriminates \emph{claims} says little about
how well it discriminates the model's \emph{own errors}, and here the second is the weaker.
The frontier arm supplies neither (Section~\ref{sec:calibration}).

\item A cascade evaluated the way a cascade should be: cross-fitted thresholds, an oracle
upper bound, a measurement of how much accuracy in-sample threshold tuning invents, and a
matched random-escalation control that the routing signal does not beat
(Section~\ref{sec:tuned}).
\end{enumerate}

We do not claim that the cascade is more accurate than the frontier model, that confidence
routing beats random routing at this sample size, or that any of this transfers beyond
grounded claim verification (Section~\ref{sec:limitations}, Appendix~\ref{app:limitations}).

\section{Related Work}
\label{sec:related}

\paragraph{Grounded claim verification.} Deciding whether a document supports a claim is the
core of automatic factuality evaluation for summarization and for retrieval-augmented
generation \citep{lewis2020rag}. Entailment-style scorers such as SummaC
\citep{laban2022summac} and the TRUE benchmark \citep{honovich2022true} established the task
shape; AggreFact \citep{tang2023aggrefact} and LLM-AggreFact with MiniCheck
\citep{tang2024minicheck} consolidated human-labelled datasets into one leaderboard and showed
that small specialized checkers can approach much larger judges. RAGTruth
\citep{niu2024ragtruth} supplies word-level hallucination annotations, while FActScore
\citep{min2023factscore} and RAGAS \citep{es2023ragas} build pipeline-level scores on
per-claim verification. We work inside that definition and change only the machine that
answers it.

\paragraph{LLM-as-judge.} Using a strong LLM to grade text is standard practice
\citep{zheng2023mtbench, liu2023geval}, along with its pathologies: position effects,
verbosity preference, and a preference for a judge's own generations
\citep{panickssery2024selfpref}. Our frontier arm is such a judge reduced to its minimal
useful form, one enumerated token, which removes the verbosity channel and keeps the cost
structure of generation.

\paragraph{Cascades and routing.} Spending little on easy inputs and escalating hard ones is old
\citep{viola2001rapid} and has been rebuilt for LLMs repeatedly. FrugalGPT
\citep{chen2023frugalgpt} chains models behind a learned scorer; Hybrid LLM
\citep{ding2024hybridllm} and RouteLLM \citep{ong2024routellm} train routers on query
features and preference data; meta-model selection \citep{sakota2024flyswat} predicts which
model suffices; and token-level uncertainty has been analysed as a deferral signal
\citep{gupta2024cascades, yue2024cascades}. Selective prediction supplies the theory, from the
reject option \citep{chow1970reject} through risk-coverage analysis
\citep{elyaniv2010selective, geifman2017selective}, and calibration the diagnostics
\citep{brier1950verification, naeini2015ece, guo2017calibration}. For LLMs, self-reported and
logprob-derived confidences are well studied \citep{kadavath2022know, tian2023justask,
xiong2024express}.

\paragraph{What is new here, stated narrowly.} Cascading, confidence-thresholded deferral and
cost-quality frontiers are all old. We claim three smaller
things. First, the cheap tier is a non-generative decision model whose probability is native
to a declared enumeration rather than derived from sampled tokens, and the escalation target
structurally cannot provide the same signal, so the router has one signal and we
characterize it instead of assuming it. Second, we measure error complementarity and
false verification between two systems that tie on the headline metric, which is what makes
routing pay and what an accuracy comparison hides. Third, we quantify the selection optimism
of in-sample threshold choice on this task \citep{varma2006bias, cawley2010selectionbias}, in
accuracy points and in dollars, on a sweep we published in sample ourselves and correct here.

\section{Problem Setup and Metrics}
\label{sec:metrics}

\paragraph{The decision.} Each example is a pair $(c, e)$ of a claim and its evidence
document, with a human gold label $y \in \{\text{\textsc{supported}}, \text{\textsc{not\_supported}}\}$. A
system returns a decision $\hat{d} \in \{\text{\textsc{supported}}, \text{\textsc{not\_supported}}\}$ and,
where it can, a probability $p$ that the claim is supported. Judgement is from the supplied
evidence alone. A claim that happens to be true in the world but is not entailed by the
document is \textsc{not\_supported}, which is the convention LLM-AggreFact uses and a
RAG guard needs.

\paragraph{Why the headline is macro-averaged.} Class balance varies sharply across the
\NDatasets{} component datasets, from RAGTruth at \SupportRAGTruthPos{} supported against
\SupportRAGTruthNeg{} unsupported to FactCheck-GPT at \SupportFactCheckGPTPos{} against
\SupportFactCheckGPTNeg{}. Per dataset $d$ we report balanced accuracy, and the headline is
its unweighted mean:

\begin{equation}
\mathrm{BAcc}_d = \tfrac{1}{2}\left(\mathrm{TPR}_d + \mathrm{TNR}_d\right),
\qquad
\mathrm{BAcc} = \frac{1}{|D|} \sum_{d \in D} \mathrm{BAcc}_d .
\end{equation}

The choice moves the number a long way. Pooling every claim into a single confusion matrix
gives \JevPooledBAcc\% for the decision model against \JevBAcc\% macro-averaged, because a
pooled sensitivity is set by whichever datasets supply most of the supported claims and a
pooled specificity by whichever supply most of the unsupported ones. We follow the
LLM-AggreFact convention and use the macro form everywhere, including inside the cascade
analysis.

\paragraph{False verification.} A verifier's two errors are not worth the same. Refusing to
verify a supported claim costs a retry, an escalation or a hedge shown to a user; verifying
an unsupported claim ships a hallucination with an approval stamp on it. We therefore report
the false-verification rate as a first-class metric:

\begin{equation}
\mathrm{FVR} = \frac{\left|\{ i : y_i = \text{\textsc{not\_supported}},\ \hat{d}_i = \text{\textsc{supported}} \}\right|}
                    {\left|\{ i : y_i = \text{\textsc{not\_supported}} \}\right|},
\end{equation}

computed over the pooled \NGoldUnsupported{} unsupported claims, together with verified
precision, the probability that a claim the system passes is genuinely supported. False
verification is the complement of specificity, which balanced accuracy averages against
sensitivity at half weight and then averages again across datasets, so two systems can land
on the same headline while differing sharply in how often they approve what the evidence
does not support. It deserves its own column rather than a derivation from two others.

\paragraph{Probability provenance.} Probabilities attached to a verdict are not
interchangeable, and comparing them without saying where they came from is a category error
the calibration literature keeps correcting. We label every probability in this
paper with one of three provenances.

\begin{description}
\item[Native.] The model emits a probability distribution over the declared decision
enumeration as part of its typed output. The decision model evaluated here is of this kind.
\item[Logprob-derived.] The probability is reconstructed from token-level log-probabilities
over the label tokens of a generative model. This requires the provider to expose logprobs.
\item[None.] The provider exposes no token probabilities, and the model can only be asked to
state a confidence in words. A verbalized self-report is not admitted into this taxonomy and
is never placed beside a native probability in a table, because the two quantities are not
estimating the same thing \citep{xiong2024express}.
\end{description}

Our frontier arm falls in the third category: the API rejects the logprobs parameter for
this model outright, with its refusal recorded in \texttt{runs/model\_live\_check.json},
so the comparison we would have liked to run is unavailable, and we
report that rather than substituting a self-reported number.

\paragraph{Discrimination over claims against discrimination over errors.} A probability can
be good at one of these and mediocre at the other, and only the second justifies routing.
The first is the AUROC of $p$ against the gold label, which asks whether the score separates
supported claims from unsupported ones. The second is the AUROC of the model's confidence,
$\max(p, 1-p)$, against its own correctness, which asks whether the score separates the
claims this model gets right from the ones it gets wrong. We call the second the selective
AUROC \citep{elyaniv2010selective, geifman2017selective}, and we report Brier score,
negative log loss and expected calibration error over equal-mass bins alongside it, with a
second binning as a sensitivity check because ECE moves with bin count
\citep{naeini2015ece, guo2017calibration}.

\paragraph{Cost.} Cost is measured token counts multiplied by a timestamped price snapshot,
at standard service tier with no caching, reported in dollars per thousand claims. Every
cost figure is therefore a modelled cost and not an invoice, and
Appendix~\ref{app:limitations} says what that does and does not support.

\section{Systems Under Test}
\label{sec:systems}

\paragraph{The decision model.} \texttt{jev-1.13.0} is a commercial non-generative model from
TypeSafe \citep{typesafe2026jev, typesafe2026models}. The request carries claim and evidence
as named state keys with a question whose answer type is a declared enumeration; the response
is one member plus a distribution over them. There is no sampling loop and no verdict to parse: response validity was
\JevValidRatePct\%, not an independent measurement, since the sample is the set of claims
an earlier pass of this model answered (Section~\ref{sec:same-score}). We take the probability of \textsc{supported} as the
native probability; the model also returns a scalar it calls confidence, which we log and
never report, because mixing two notions of certainty makes calibration comparisons
meaningless. The version is pinned on every call, the SDK default being a moving alias
\citep{typesafe2026citationcheck}.

\paragraph{The frontier model.} \texttt{gpt-6-astra} is run as an engineer would deploy it
for a binary classifier: a short prompt, an enumerated response mode, no tools,
provider-default temperature and a tight completion budget. Response validity was
\AstraValidRatePct\%. The provider reports no token log-probabilities for this model, so
its probability provenance is \emph{none} and it supplies no routing signal. We run two
reasoning efforts, treating low as primary: what a cost-conscious team picks.

The provider returns a per-call count of internal reasoning tokens, and we print what came
back rather than what we asked for: at low effort the median is \AstraReasoningMedian\ with
\AstraReasoningZeroN\ of \NPaired\ calls reporting none, at high
\AstraHighReasoningMedian\ with \AstraHighReasoningZeroN. The obvious objection is that our frontier arm was
told not to think and duly did not; these counts are the honest form of it. Both efforts run under one prompt, byte for byte, ending by forbidding reasoning in the
output (Appendix~\ref{app:prompts}). They do not run under one completion-token cap:
\AstraTokenCap\ against \AstraHighTokenCap, because the cap covers the reasoning too and
the high arm's longest reply alone reaches \AstraHighMaxOutputTokens\ tokens, so the low
cap would have truncated it, the failure Appendix~\ref{app:incidents} records. Raising the effort parameter moved the score
by \AstraAstraHighDeltaPP\ points at exact McNemar $p = \AstraAstraHighMcNemarP$, so the
contrast bounds what the knob buys together with the headroom it needs, under our
instruction, rather than testing a prompt that invites reasoning. That configuration was
never run and Section~\ref{sec:limitations} lists it as missing.

\paragraph{Prompt parity, including the fix we caught ourselves.} Both systems receive the same
semantic instructions: judge only from the evidence, and mark a claim unsupported if any
materially important part of it is missing, contradicted, numerically or temporally wrong,
over-scoped or embellished. A pre-publication audit found the decision model's
criteria naming wrong entities as grounds for \textsc{not\_supported} while the frontier
prompt did not, an asymmetry favouring the decision model, since entity-swap hallucinations
are a large share of the AggreFact negatives. We added the trigger and re-ran both frontier
configurations; every frontier number here comes from the corrected runs, the pre-correction
runs are retained so the fix stays auditable, and both prompts are in
Appendix~\ref{app:prompts}.

\paragraph{Token accounting.} The decision model's requests are the longer,
\JevMeanInputTokens{} input tokens on average against \AstraMeanInputTokens{}, so its cost
advantage is not from sending less. Nor is it, to any degree worth reporting, from its vendor
charging for input only: re-pricing those output tokens at the vendor's own input rate
raises the cost per thousand from \$\JevCostPerK{} to \$\JevCostPerKIfOutputBilled{} and
the ratio from \CostRatioAstraOverJev$\times$ only to \CostRatioIfOutputBilled$\times$,
so that policy accounts for \FreeOutputShareOfRatioPct\% of it and no more. The ratio is
the product of three measured factors and we print all three: input price differs by
\InputPriceRatio-fold, \$\AstraInputPricePerMtok{} per million against
\$\JevInputPricePerMtok{}; dividing by \InputTokenRatioJevOverAstra, the decision model
sending the longer request, pulls that to \RatioFromPriceAndTokens; and the frontier arm's
output tokens at \$\AstraOutputPricePerMtok{} per million multiply it by
\AstraOutputUplift. The identity is asserted in the analysis script, so a change breaking it
fails the run rather than reaching this page.

\begin{table}[t]
\centering
\small
\begin{tabular}{llccc}
\toprule
Configuration & Response & Prob.\ provenance & Mean in / out tokens & Cost / 1k \\
\midrule
\texttt{jev-1.13.0}               & typed enum value, no text & native & \JevMeanInputTokens\ / \JevMeanOutputTokens       & \$\JevCostPerK \\
\texttt{gpt-6-astra}, effort low  & enumerated token          & none   & \AstraMeanInputTokens\ / \AstraMeanOutputTokens   & \$\AstraCostPerK \\
\texttt{gpt-6-astra}, effort high & enumerated token          & none   & \AstraHighMeanInputTokens\ / \AstraHighMeanOutputTokens & \$\AstraHighCostPerK \\
\bottomrule
\end{tabular}
\caption{The three configurations. Token counts are per-request means measured on the
\NPaired{} paired claims, and costs apply the uncached standard-tier price snapshot to those
counts. The decision model returns no text and its provider nonetheless reports more output
tokens for it than the model that writes one, which the spread of the counts settles rather
than the means:
over all \NPaired{} claims its output length takes
\JevDistinctOutputTokenCounts{} values, \JevMinOutputTokens{} or \JevMaxOutputTokens, so what
those tokens measure is a fixed serialisation envelope wrapping the enumerated value and its
probability, not prose. The frontier arm's length varies with the claim, over
\AstraDistinctOutputTokenCounts{} values from \AstraMinOutputTokens{} to
\AstraMaxOutputTokens{} at standard effort and \AstraHighDistinctOutputTokenCounts{} from
\AstraHighMinOutputTokens{} to \AstraHighMaxOutputTokens{} at high effort. The decision
model's output tokens are billed at zero by its provider. A fourth
configuration, the open-weight checker MiniCheck-Flan-T5-Large \citep{tang2024minicheck},
was run on the same claims and then excluded, because our port of it scored far below every
row here and we could not attribute the gap to the model rather than to the port;
Appendix~\ref{app:limitations} gives the evidence and the numbers.}
\label{tab:systems}
\end{table}

\section{Experimental Setup}
\label{sec:setup}

\paragraph{Data.} We use the development split of LLM-AggreFact \citep{tang2024minicheck} at
a pinned revision, recorded with its file hash in the run manifest. The evaluation
sample is \NPaired{} claims, in equal numbers from each of the \NDatasets{} component
datasets, so no dataset dominates the macro average through size alone. Its example-ID
list is released and hashed with the predictions but regenerated by no code here, so it
ships as data rather than as a reproducible draw. Gold labels are \NGoldSupported{} supported and \NGoldUnsupported{}
unsupported, a supported rate of \GoldSupportedRatePct\%. Both classes are present in every
dataset, so macro-balanced accuracy is defined everywhere, though the thinnest minority
class is \ThinnestMinorityClassN{} claims, which is why the cross-fitting in
Section~\ref{sec:cascade} scores assembled out-of-fold predictions rather than folds.

\paragraph{Split discipline.} Prompts, thresholds and every cascade parameter were chosen on
this development split, one threshold out of fold within it and the rest in sample. The test
split was never read: the loader refuses to fetch it until the protocol tag is pushed, a
mechanical guard rather than a promise. Our intervals are therefore
development-split intervals, and neither system can be assumed uncontaminated by a benchmark
public for years.

\begin{table}[t]
\centering
\small
\begin{tabular}{lccc}
\toprule
Dataset & $n$ & Supported & Unsupported \\
\midrule
AggreFact-CNN   & \SupportAggreFactCNNN    & \SupportAggreFactCNNPos    & \SupportAggreFactCNNNeg \\
AggreFact-XSum  & \SupportAggreFactXSumN   & \SupportAggreFactXSumPos   & \SupportAggreFactXSumNeg \\
ClaimVerify     & \SupportClaimVerifyN     & \SupportClaimVerifyPos     & \SupportClaimVerifyNeg \\
ExpertQA        & \SupportExpertQAN        & \SupportExpertQAPos        & \SupportExpertQANeg \\
FactCheck-GPT   & \SupportFactCheckGPTN    & \SupportFactCheckGPTPos    & \SupportFactCheckGPTNeg \\
Lfqa            & \SupportLfqaN            & \SupportLfqaPos            & \SupportLfqaNeg \\
RAGTruth        & \SupportRAGTruthN        & \SupportRAGTruthPos        & \SupportRAGTruthNeg \\
Reveal          & \SupportRevealN          & \SupportRevealPos          & \SupportRevealNeg \\
TofuEval-MediaS & \SupportTofuEvalMediaSN  & \SupportTofuEvalMediaSPos  & \SupportTofuEvalMediaSNeg \\
TofuEval-MeetB  & \SupportTofuEvalMeetBN   & \SupportTofuEvalMeetBPos   & \SupportTofuEvalMeetBNeg \\
Wice            & \SupportWiceN            & \SupportWicePos            & \SupportWiceNeg \\
\midrule
Total           & \NPaired                 & \NGoldSupported            & \NGoldUnsupported \\
\bottomrule
\end{tabular}
\caption{Composition of the evaluation sample: equal strata across the \NDatasets{}
component datasets of LLM-AggreFact, with gold class counts as labelled by the original
annotators. Per-dataset results are in Appendix~\ref{app:per-dataset}.}
\label{tab:sample}
\end{table}

\paragraph{The paired design, and what it does not buy here.} Each claim goes to every
configuration, so the systems are compared on identical inputs and their difference never
crosses the gap between two independently drawn samples
\citep{dietterich1998tests}. That removes the between-sample variance without narrowing
the difference interval, and the cause is the metric: macro-balanced accuracy averages
\NDatasets{} small per-dataset accuracies, each
contributing noise to the difference as well as to the levels, so the interval on the
difference is no narrower than either system's own (Table~\ref{tab:headline}).

\paragraph{Inference, and when we call a difference real.} The difference in macro-balanced
accuracy is estimated by paired stratified bootstrap with \BootstrapResamples{} resamples
\citep{efron1993bootstrap}, stratified by component dataset, resampling claims rather than
model outputs. Discordance is tested with an exact McNemar test \citep{mcnemar1947note} on
per-item correctness. The two do not always measure the same thing. On every
false-verification comparison they coincide, the rate being pooled
over the gold-unsupported claims and the test running on those same claims, so we require
both: a comparison separates from noise when the 95\% bootstrap interval excludes zero
\emph{and} the exact McNemar returns $p < 0.05$. On macro-balanced accuracy they do not, the
interval being on the macro average and the test on pooled per-item correctness, so there
the interval governs and the $p$-value corroborates rather than conditions. Where one
instrument clears and the other does not we say which and call the result marginal. Every
$p$-value is also Holm-Bonferroni adjusted over the \MultiplicityFamilyN\ exact McNemar tests
this paper reports, raw and adjusted side by side in Appendix~\ref{app:limitations}, and a
separation is demonstrated only where it survives. The family is fixed in code before any
adjustment, because at this size it decides an answer: adjusted,
\MultiplicitySurvivingN\ comparisons among the reported configurations separate, both on
the guard metric, while the two
tests against the excluded MiniCheck arm clear it as evidence for that exclusion.
\MultiplicityMedianMembersN\ of its \MultiplicityFamilyN\ members are medians over
cross-fitting repeats rather than single exact tests, so the adjustment is descriptive
there and claims no error rate.

\paragraph{Cost, failures, and the absence of latency.} Costs are the token counts each
provider returned, priced from \texttt{configs/} \texttt{pricing\_2026-09-20.yaml}, a
snapshot archived and hashed the day before the runs, standard tier, caching disabled. We report no latency comparison: both bulk runs used
client-side concurrency and the controlled single-stream protocol was never run.
Operational failures such as timeouts and rate limits are retried; a model-caused failure, a
malformed response, refusal or persistent provider error, is scored wrong rather than
dropped, so a system cannot improve its accuracy by declining hard cases. No configuration
here had one, and Appendix~\ref{app:limitations} says for which arms that is independent.

\paragraph{Cascade simulation.} The cascade is simulated offline from the locked prediction
files, with no additional inference: a claim is accepted from the decision
model when its probability is confident either way and otherwise takes the frontier model's
stored verdict. Thresholds are selected by cross-fitting,
\CrossfitRepeats{} repeats of \CrossfitFolds-fold stratified cross-validation, giving
\CrossfitThresholdFits{} threshold fits, every claim scored under a threshold chosen
without reference to its own fold. Two controls accompany it: escalating the same fraction
of traffic at random, which isolates what the signal contributes over the budget itself,
and an oracle escalating exactly the claims the decision model gets wrong.

\paragraph{Reproducibility and number discipline.} Each prediction file was committed when its
run finished and has not changed since. Their SHA-256 digests sit in an evidence file covered
by an OpenTimestamps proof, both released, so a reader can confirm without our cooperation
that the predictions were fixed before the analysis reading them. That is what a hash
under a timestamp establishes, and all we claim from it. The analysis
regenerates one numbers file, from which every value here is substituted at build time;
Appendix~\ref{app:repro} gives the commands.

%% file: same-score.tex

\section{Results: same score}
\label{sec:same-score}

On the \NPaired{} paired claims of the development split, \texttt{jev-1.13.0} reaches
\JevBAccTwoDP{} macro-balanced accuracy and GPT-6 Astra reaches \AstraBAccTwoDP{}. The paired
difference, on Jev minus Astra, is $\JevAstraDeltaSignedTwoDP$ points with a 95\% interval of
[$\JevAstraDeltaCILo$, $\JevAstraDeltaCIHi$] from \BootstrapResamples{} stratified paired
bootstrap resamples, and an exact McNemar test on the per-item decisions returns
$p = \JevAstraMcNemarP$. Both systems produced a parseable verdict on every claim, so no part of the comparison
rests on format failures or refusals. The two runs cost \$\JevCostPerK{} and \$\AstraCostPerK{} per
thousand claims, a factor of \CostRatioAstraOverJev{}.

\subsection{The paired difference}
\label{subsec:paired-difference}

Table~\ref{tab:paired-contrasts} gives the three contrasts between the configurations
under test, all signed first system minus second. Pairing is what makes the difference estimable at this sample size: the bootstrap resamples
the pair rather than the two runs independently (Section~\ref{sec:setup}). It does not make
the difference small, for the reason that section gives. The McNemar column tests raw
per-item accuracy pooled across datasets rather than the macro quantity in the first column,
since an exact binomial test on discordant pairs needs one binary outcome per item. It rests
on the \DisagreementPct\% of claims where the two verdicts differ, and those pairs very
nearly split evenly; Section~\ref{sec:different-errors} opens them up.
The strongest reason to hold this difference loosely runs against us. The sample is the set
of claims an earlier exploratory pass of the decision model answered, and on those claims
that pass scores \JevRerunBAccTwoDP\ against \AstraBAccTwoDP: a deficit, not a tie. We
report a later pass over the identical IDs, \JevBAccTwoDP, \JevRerunDeltaPP\ points above
it and still short of \AstraBAccTwoDP, because the exploratory one could not be
provenanced. The decision model therefore moves further between two passes of itself than
the gap this section is about; Appendix~\ref{app:limitations} gives both runs.

Raising Astra's reasoning effort does not separate the systems either. At high effort
Astra reaches \AstraHighBAccTwoDP{}, which on the table's convention puts the
Astra-against-high-effort contrast at $\AstraAstraHighDeltaSignedPP$ points with an
interval of [$\AstraAstraHighDeltaCILo$, $\AstraAstraHighDeltaCIHi$] and McNemar
$p = \AstraAstraHighMcNemarP$ on \AstraAstraHighMcNemarStat{} discordant pairs, while its
cost rises to \$\AstraHighCostPerK{} per thousand claims, or \CostRatioAstraHighOverJev{}
times Jev's. Jev against that configuration is $\JevAstraHighDeltaSignedPP$ points,
[$\JevAstraHighDeltaCILo$, $\JevAstraHighDeltaCIHi$], $p = \JevAstraHighMcNemarP$.
Whatever is holding both systems where they sit, more deliberation at the frontier end does
not move it.

\begin{table}[t]
\centering\small
\caption{Paired contrasts on the \NPaired{} development claims. $\Delta$ is macro-balanced
accuracy, first system minus second, in percentage points, with a 95\% interval from
\BootstrapResamples{} stratified paired bootstrap resamples. The McNemar column is an exact
test on raw per-item accuracy pooled over datasets, and its statistic is the smaller of the
two discordant counts.}
\label{tab:paired-contrasts}
\begin{tabular}{lrrrr}
\toprule
Contrast & $\Delta$ (pp) & 95\% CI & McNemar stat. & $p$ \\
\midrule
Jev vs.\ Astra                & $\JevAstraDeltaSignedPP$      & [$\JevAstraDeltaCILo$, $\JevAstraDeltaCIHi$]           & \JevAstraMcNemarStat      & \JevAstraMcNemarP \\
Jev vs.\ Astra (high effort)  & $\JevAstraHighDeltaSignedPP$  & [$\JevAstraHighDeltaCILo$, $\JevAstraHighDeltaCIHi$]   & \JevAstraHighMcNemarStat  & \JevAstraHighMcNemarP \\
Astra vs.\ Astra (high effort)& $\AstraAstraHighDeltaSignedPP$& [$\AstraAstraHighDeltaCILo$, $\AstraAstraHighDeltaCIHi$]& \AstraAstraHighMcNemarStat& \AstraAstraHighMcNemarP \\
\bottomrule
\end{tabular}
\end{table}

\subsection{The ordering changes with the metric}
\label{subsec:metric-flip}

On raw accuracy the ordering reverses and Jev leads, \JevRawAcc\% against \AstraRawAcc\%.
The cause is class prevalence. Gold labels here are \GoldSupportedRatePct\% supported, Jev
calls \JevPredSupportedRate\% supported and Astra \AstraPredSupportedRate\%, so the more
permissive system harvests more correct answers from the larger class and pays on the
smaller one, a trade raw accuracy rewards at this prevalence and balanced accuracy does
not. Pooling the \NDatasets{} datasets into one
confusion matrix before balancing puts Astra back in front, \AstraPooledBAcc{} against
\JevPooledBAcc{}. Macro $F_1$ favours Jev, \JevMacroFOne{} against \AstraMacroFOne{}, and
the Matthews correlation coefficients differ only in the third decimal, \JevMCC{} against
\AstraMCC{}. Table~\ref{tab:metric-flip} collects them: reasonable summary metrics,
disagreeing about the winner, over one set of predictions. Macro-balanced accuracy is the
headline for the reasons given in Section~\ref{sec:metrics}: it is the convention LLM-AggreFact established for this
task \citep{tang2024minicheck}, it refuses to let a verifier buy its score from the majority
class, and a component deployed as a guard is answerable for both error directions. A
different metric would have handed a reader a different verdict.

\begin{table}[t]
\centering\small
\caption{Summary metrics on one set of predictions over the same \NPaired{} claims.
All values are percentages except MCC. The macro-balanced accuracy row is printed to two
decimals because Table~\ref{tab:paired-contrasts} prints the difference of those two cells
and a reader is entitled to reproduce it; the other rows are differenced nowhere.
Gold labels are \GoldSupportedRatePct\% supported,
which is what the raw-accuracy row is responding to.}
\label{tab:metric-flip}
\begin{tabular}{lrrl}
\toprule
Metric & Jev & GPT-6 Astra & Leads \\
\midrule
Macro-balanced accuracy  & \JevBAccTwoDP  & \AstraBAccTwoDP  & Astra \\
Pooled balanced accuracy & \JevPooledBAcc & \AstraPooledBAcc & Astra \\
Raw accuracy             & \JevRawAcc     & \AstraRawAcc     & Jev \\
Macro $F_1$              & \JevMacroFOne  & \AstraMacroFOne  & Jev \\
MCC                      & \JevMCC        & \AstraMCC        & Jev \\
\midrule
Claims called supported  & \JevPredSupportedRate & \AstraPredSupportedRate & \\
\bottomrule
\end{tabular}
\end{table}

\subsection{What the interval does not support}
\label{subsec:not-equivalence}

The interval runs from $\JevAstraDeltaCILo$ to $\JevAstraDeltaCIHi$ points, so these data
are compatible with a gap of several points in either direction, and a gap at either end
would be a large operational difference for a component that adjudicates claims at volume.
We therefore claim no equivalence: no margin was fixed in advance and no
two-one-sided-tests procedure was run, so Table~\ref{tab:paired-contrasts} reports a
failure to detect a difference rather than evidence that none exists. The resolution limit
is structural. Macro-balanced accuracy averages \NDatasets{} small datasets, the thinnest
minority class among them holds \ThinnestMinorityClassN{} claims, and one flipped decision
inside that class moves the macro average by more than the whole gap between the two
systems.

What survives is narrow and worth having: two systems separated by \CostRatioAstraOverJev{}
times in price differ by \JevAstraDeltaPP{} points on the metric this task is
conventionally scored by, and the cheaper emits no text at all. The same
\NPaired{} claims that produce that tie also record which claims each system got wrong, and
those records do not match.

%% file: different-errors.tex

\section{Results: different errors}
\label{sec:different-errors}

Two systems that score the same can still be wrong about different things, and these two are.
Jev and GPT-6 Astra agree on \AgreementPct\% of the \NPaired\ paired claims and disagree on
\DisagreementPct\%, and against the human labels that disagreement splits close to evenly in
both directions, the fact the rest of this paper is built on.

\subsection{The disagreement splits both ways}

\begin{figure}[t]
  \centering
  \includegraphics[width=0.62\linewidth]{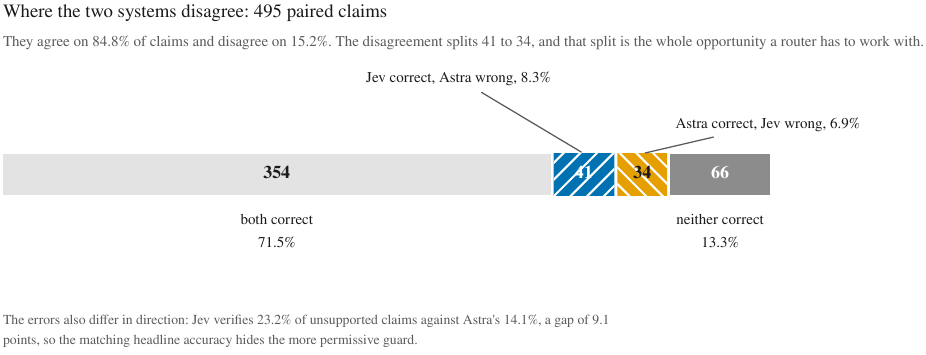}
  \caption{Error complementarity on the paired dev set. The agreement block splits into
  \BothCorrect\ claims both systems get right and \NeitherCorrect\ neither does; the
  disagreement block splits \JevOnlyCorrect\ to \AstraOnlyCorrect. Percentages are of
  \NPaired\ claims.}
  \label{fig:complementarity}
\end{figure}

Jev is correct where Astra is wrong on \JevOnlyCorrect\ claims, and Astra is correct
where Jev is wrong on \AstraOnlyCorrect. Table~\ref{tab:complementarity} gives the full
decomposition and Figure~\ref{fig:complementarity} draws it. The exact McNemar test of
Section~\ref{sec:same-score} is computed on exactly these two discordant cells and returns
$p = \JevAstraMcNemarP$. A test that cannot separate \JevOnlyCorrect\ from
\AstraOnlyCorrect\ does not say the two systems make the same mistakes, and a
reader taking the tie as interchangeable behaviour inherits an error profile nobody measured.
The \NeitherCorrect\ claims neither gets right put a floor under anything assembled from
this pair.

\begin{table}[t]
  \centering
  \caption{Paired outcomes on the \NPaired\ dev claims, scored against the human labels.
  The two discordant cells are what any composition of the two systems has to work with;
  the \NeitherCorrect\ claims in the fourth row are beyond both systems.}
  \label{tab:complementarity}
  \begin{tabular}{lr}
    \toprule
    Outcome & Claims \\
    \midrule
    Both correct                 & \BothCorrect \\
    Jev correct, Astra wrong     & \JevOnlyCorrect \\
    Astra correct, Jev wrong     & \AstraOnlyCorrect \\
    Neither correct              & \NeitherCorrect \\
    \midrule
    Total                        & \NPaired \\
    \bottomrule
  \end{tabular}
\end{table}

\subsection{The cheaper system is the more permissive guard}

Jev verifies \JevFVR\% of the claims the human annotators marked unsupported, against
\AstraFVR\% for Astra: a gap of \FVRGapPP\ points over the \NGoldUnsupported{}
gold-unsupported claims in the sample. That gap gets the accuracy gap's treatment from
Section~\ref{sec:same-score} and unlike it survives. Among those \NGoldUnsupported\ claims,
\FVRBothFalseVerify\ are waved through by both systems,
\FVRJevOnlyFalseVerifies\ by Jev alone and \FVRAstraOnlyFalseVerifies\ by Astra alone. An
exact McNemar on those discordant claims returns $p = \FVRGapMcNemarP$, $\FVRGapMcNemarPHolm$
adjusted, and a paired stratified bootstrap over the gold-unsupported claims puts the gap at
[\FVRGapCILo, \FVRGapCIHi] points, so both instruments clear the criterion of
Section~\ref{sec:setup} on either reading. The same \NPaired\ claims that leave the headline
difference inside the noise separate this one cleanly. The asymmetry shows in verified
precision too, the share of claims a system calls supported that really are:
\JevVerifiedPrecision\% for Jev, \AstraVerifiedPrecision\% for Astra. A guard buys exactly
that quantity, and at matched macro-balanced accuracy the system cheaper by
\CostRatioAstraOverJev{} lets through more of what it exists to stop.

The mechanism is how often each says supported at all. Gold support runs at
\GoldSupportedRatePct\% of claims, Jev predicts supported on \JevPredSupportedRate\% and
Astra on \AstraPredSupportedRate\%, so Astra sits well below the base rate and Jev just
under it.
That is also the whole of the raw-accuracy flip in Section~\ref{sec:same-score}: Jev's
\JevRawAcc\% against Astra's \AstraRawAcc\% is earned on the majority class and paid for on
the class a guard cares about. Section~\ref{sec:calibration} shows Jev's false-verification
rate is tunable by gating on its own probability, a lever Astra does not have.

\subsection{Per dataset, the advantage changes hands}

Across the \NDatasets\ constituent datasets of LLM-AggreFact \citep{tang2024minicheck},
Jev scores higher on \DatasetsJevHigher\ and Astra on \DatasetsAstraHigher.
Figure~\ref{fig:per-dataset} plots the pairs on a full zero-to-one-hundred axis, and
Table~\ref{tab:per-dataset} in Appendix~\ref{app:per-dataset} gives every cell with its
class support. The two widest gaps run in opposite directions: Jev reaches
\JevBAccExpertQA\ on ExpertQA where Astra reaches \AstraBAccExpertQA, below the coin-flip
line and the worst single cell either system produces, while on AggreFact-CNN the order
reverses, \JevBAccAggreFactCNN\ against \AstraBAccAggreFactCNN. Those two cells do largely
cancel, and saying so settles nothing, because the other \NRemainingDatasets\ do the same
(Appendix~\ref{app:per-dataset}, cell by cell): the honest description is not that
\NRemainingDatasets\ datasets agree and two disagree. All \NDatasets\ disagree, in both
directions, by \MeanAbsRemainingPP\ points on average away from the two extremes and by far
more at them: \SumAbsDeltasPP\ points of per-dataset disagreement cancel down to a signed residue of ${\SignedSumDeltasPP}$ points, and the
\JevAstraDeltaPP-point headline is that residue spread across the \NDatasets\ datasets, each
about \SupportAggreFactCNNN\ claims. That is fragile: no single dataset should be
quoted alone, and the aggregate is not evidence these two systems behave alike anywhere.

\begin{figure}[t]
  \centering
  \includegraphics[width=0.62\linewidth]{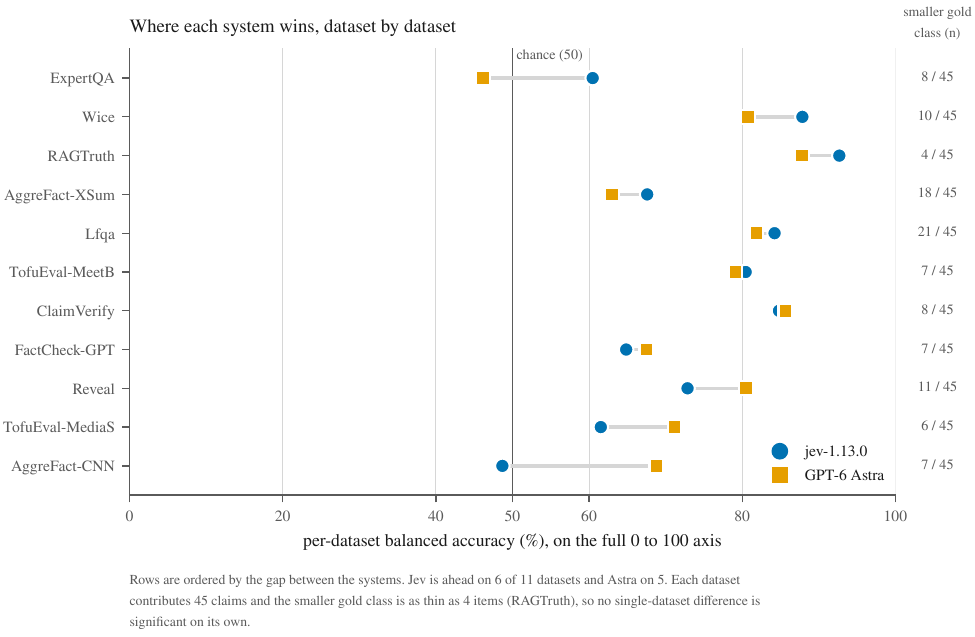}
  \caption{Per-dataset balanced accuracy, Jev against Astra, on a full
  zero-to-one-hundred axis. The advantage changes hands \DatasetsJevHigher\ times to
  \DatasetsAstraHigher. Each row is one dataset of about \SupportAggreFactCNNN\ claims,
  so the rank order of the rows is the least stable result in this paper.}
  \label{fig:per-dataset}
\end{figure}

\subsection{What RAGTruth cannot tell us}

Jev also leads on RAGTruth \citep{niu2024ragtruth}, \JevBAccRAGTruth\ against
\AstraBAccRAGTruth, and that lead carries none of the permissiveness finding above. Our
sample draws \SupportRAGTruthN\ RAGTruth claims, \SupportRAGTruthNeg\ of them
gold-unsupported, and both systems label all \SupportRAGTruthNeg\ correctly, which the
reported scores force: at a minority class that thin, one miss there would drop the dataset
below the balanced accuracy either records, so both specificities sit at the top of the
scale and the gap comes from the supported side. They are not pinned there: the earlier pass of the decision
model false-verifies \JevRerunFalseVerifiedRAGTruth\ of those \SupportRAGTruthNeg\ claims
and takes the dataset to \JevRerunBAccRAGTruth, so the ceiling is a property of this run
rather than of either system. The dataset whose name most invites a reading about
hallucination catching says nothing about it. The caution generalises: the thinnest unsupported class holds
\ThinnestMinorityClassN\ claims and several others are in single digits, so we read the
false-verification gap only at the aggregate over all \NGoldUnsupported.

\subsection{Errors that differ can be composed}

Two systems whose errors coincide are substitutes, and the only question is which is cheaper.
These two are not that, though they are far from independent: their per-item correctness
agrees at Cohen's $\kappa = \ErrorKappa$ and they fail together on \NeitherCorrect\ claims
where independence predicts \NeitherExpectedIndependent, a factor of \JointFailureExcess.
Composition needs not independent errors but discordant cells large enough to route between,
and those the table shows: \JevOnlyCorrect\ claims where paying \CostRatioAstraOverJev\ times
more buys a worse answer, \AstraOnlyCorrect\ where it buys a better one, and a false-verification gap running
against the direction the headline scores suggest. A system routing each claim to whichever
is right about it would beat both, by a margin Section~\ref{sec:cascade} measures against an
oracle. Doing that in production needs a signal, available before the expensive call, for
which claims the cheap system will get wrong. Whether either supplies one is
Section~\ref{sec:calibration}.

%% file: cascade.tex
%

\section{Results: calibration, and who even has a probability}
\label{sec:calibration}

Jev's native probability separates supported claims from unsupported ones well, and its
own correct verdicts from its own mistakes only moderately. Scored
against the gold label, $p_{\mathrm{supported}}$ reaches an AUROC of \JevAUROC\ with
AUPRC \JevAUPRC. Scored against whether Jev itself was right, the confidence
$c=\max(p,\,1-p)$ reaches a selective AUROC of \JevSelectiveAUROC\ with selective AUPRC
\JevSelectiveAUPRC. Both come from the same \NPaired\ probabilities, answer different
questions, and every routing decision here turns on the second.

A verifier can rank claims well and still be confidently wrong on a recognisable family of
them, the items a router must catch, since it sees the confidence and never the label.
Quoting \JevAUROC\ beside a routing claim overstates what a router has.

\subsection{Calibration of the native probability}

Jev's probability is under-confident across the lower half of the scale and over-confident
in a narrow band near the top. The Brier score is \JevBrier\ against the \JevBrierConstantHalf\ a
constant one-half forecast scores on these labels, and the expected calibration error is
\JevECE\ over ten equal-frequency bins, moving to \JevECESensitivity\ at fifteen. The
negative log likelihood is \JevNLL, and we read no verdict off it:
\JevNLLClipSharePct\% of it is the \JevNLLClipN\ claims Jev called certain and got wrong,
charged at
a clipping constant \texttt{configs/benchmark.yaml} marks provisional. Certain there is the
vendor's resolution, not the model's conviction: $p_{\mathrm{supported}}$ arrives at
\JevProbDecimals\ decimals, a reported zero means below \JevProbHalfStep, and what to
charge it is ours; at a constant that size the same predictions give
\JevNLLAtHalfStepClip\ (Appendix~\ref{app:limitations}).

\begin{figure}[t]
\centering
\includegraphics[width=0.62\linewidth]{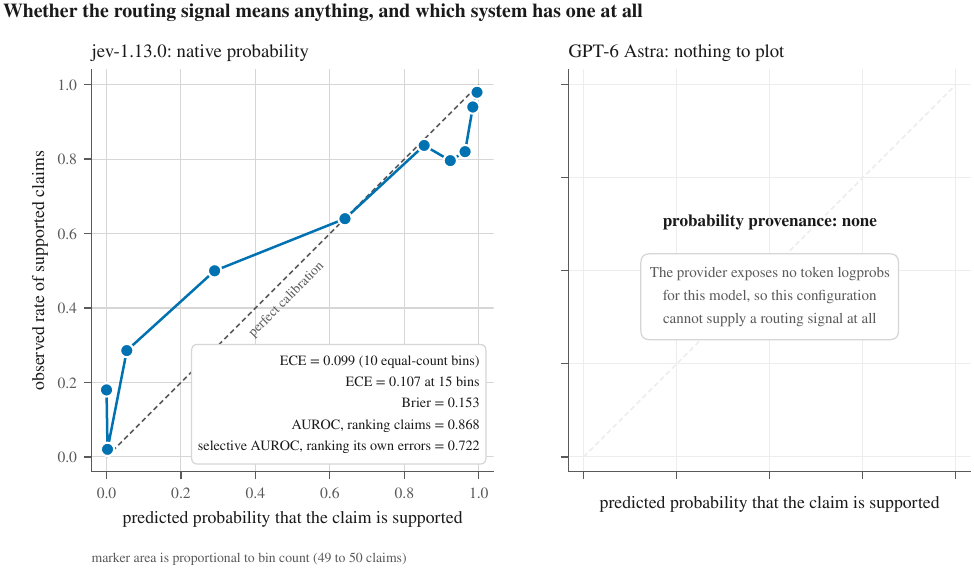}
\caption{Reliability of Jev's native probability over ten equal-frequency bins, with ECE
annotated. Equal-frequency binning follows the protocol rather than the equal-width
convention of \citet{guo2017calibration}, because this probability piles up at zero and
one, leaving most equal-width bins empty. The Astra panel states that no probability
exists for that configuration rather than showing an empty axis: the provider returns no
token logprobs for this model class.}
\label{fig:reliability}
\end{figure}

Figure~\ref{fig:reliability} shows where. The curve sits above the diagonal at and below
\JevCalUnderConfidentAtOrBelow, widest at a predicted \JevCalWorstP\ observed
\JevCalWorstObserved, so claims it calls unlikely are supported far more often than it
says. It is at or under the diagonal at \JevCalBinsBelowDiagonalN\ of ten bins, from
\JevCalFirstAtOrBelowP\ up, clearing the binning tolerance between
\JevCalOverConfidentLo\ and \JevCalOverConfidentHi.
Interior bins, between zero and \JevCalInteriorHi, carry \JevCalInteriorShareOfECEPct\% of
the error; but that widest gap, at a predicted \JevCalWorstP, is a confidence of
\JevCalWorstConfidence, the quantity a router sees, and the flattest bin is the least
confident, at \JevCalBestP. The
gap at a reported zero is equal-count binning cutting the \JevProbAtZeroN\ predictions
reported as exactly zero, \JevCalZeroInLowestBinN\ into the lowest bin. A probability
furthest off where a router trusts it is the modest selective AUROC from another angle,
bounding what any threshold rule can recover.

\subsection{Which systems have a probability at all}

Astra reports no probability, at either reasoning budget: asked for token logprobs the
provider returns a 400 for this model class, recorded verbatim in
\texttt{runs/model\_live\_check.json}, so the escalation target in our cascade supplies
neither a native probability nor a logprob-derived one. That file carries no such probe of the
Claude-family model in our registry, only our own entry marking it as having none, so that
one is a registry decision, not a measurement. The 400 is a property of the deployed
interface rather than a gap in ours, and it decides the architecture: a frontier-first
cascade, where the expensive system answers and escalates what it is unsure about, cannot be
built with these models, so the only direction is cheap-first. Any confidence attached to Astra's verdicts
would be a self-reported verbal estimate, a different object from a trained probability
head and known to behave poorly under distribution shift \citep{kadavath2022know}. We
substitute neither for the other.

\subsection{What the probability buys before any routing}

Gating on Jev's confidence alone, with no second system, buys precision with
coverage: verified precision rises and false verification falls at every step of
the grid (Table~\ref{tab:gating}). Of the claims Jev calls supported at full coverage,
\GateAtFiftyVerifiedPrecision\% really are, at a false-verification rate of
\GateAtFiftyFVR\%. At the tightest gate it answers \GateAtNinetyNineCoverage\% of claims
with verified precision \GateAtNinetyNineVerifiedPrecision\% and false verification
\GateAtNinetyNineFVR\%.

The balanced accuracy in that last row is not the same quantity as the ones above it, for
the reason the Datasets column and caption give: the datasets dropping out of its
average are AggreFact-CNN and TofuEval-MediaS, where Jev scores \JevBAccAggreFactCNN\ and
\JevBAccTofuEvalMediaS\ at full coverage, two of its three weakest cells, so the row gains
from the changed denominator as well as from the gate.

\begin{table}[t]
\centering\small
\caption{Gating on Jev's native confidence, Jev alone. Rows are the confidence thresholds
of the sweep grid, the \texttt{THRESHOLDS} constant in
\texttt{scripts/verify/paired\_analysis.py}, indexed by the coverage each one produces.
The grid is not preregistered, and no analysis decision in this paper is
(Section~\ref{sec:limitations}). Balanced accuracy, verified precision and false
verification are computed on the covered claims only. Datasets is how many component
datasets the macro on that row is taken over: at the tightest gate
\GateAtNinetyNineNDatasetsDropped\ of them retain no claim of one gold class, so
\GateAtNinetyNineBAcc\ is a macro over \GateAtNinetyNineNDatasets\ and not over
\NDatasets.}
\label{tab:gating}
\begin{tabular}{rrrrr}
\toprule
Coverage (\%) & Datasets & Macro BAcc & Verified prec.\ (\%) & FVR (\%) \\
\midrule
\GateAtFiftyCoverage        & \GateAtFiftyNDatasets        & \GateAtFiftyBAcc        & \GateAtFiftyVerifiedPrecision        & \GateAtFiftyFVR \\
\GateAtSeventyCoverage      & \GateAtSeventyNDatasets      & \GateAtSeventyBAcc      & \GateAtSeventyVerifiedPrecision      & \GateAtSeventyFVR \\
\GateAtEightyCoverage       & \GateAtEightyNDatasets       & \GateAtEightyBAcc       & \GateAtEightyVerifiedPrecision       & \GateAtEightyFVR \\
\GateAtNinetyCoverage       & \GateAtNinetyNDatasets       & \GateAtNinetyBAcc       & \GateAtNinetyVerifiedPrecision       & \GateAtNinetyFVR \\
\GateAtNinetyFiveCoverage   & \GateAtNinetyFiveNDatasets   & \GateAtNinetyFiveBAcc   & \GateAtNinetyFiveVerifiedPrecision   & \GateAtNinetyFiveFVR \\
\GateAtNinetySevenCoverage  & \GateAtNinetySevenNDatasets  & \GateAtNinetySevenBAcc  & \GateAtNinetySevenVerifiedPrecision  & \GateAtNinetySevenFVR \\
\GateAtNinetyNineCoverage   & \GateAtNinetyNineNDatasets   & \GateAtNinetyNineBAcc   & \GateAtNinetyNineVerifiedPrecision   & \GateAtNinetyNineFVR \\
\bottomrule
\end{tabular}
\end{table}

Balanced accuracy over the covered claims does not rise monotonically: the row at
\GateAtNinetyCoverage\% coverage scores \GateAtNinetyBAcc, below the
\GateAtEightyBAcc\ of the looser row above it. Each row is a different subsample and single
operating points move by more than a point at this size; Section~\ref{sec:tuned} measures
that movement rather than describing it. Gating also leaves the abstained claims unanswered, and a
pipeline still has to answer them.

\section{Confidence-gated cascading}
\label{sec:cascade}

Answering the confident claims with Jev and escalating the rest to Astra reaches
\CascadeCrossfitBAccTwoDP\ macro-balanced accuracy at \$\CascadeCrossfitCost\ per thousand
claims, threshold chosen out of fold, against \AstraBAccTwoDP\ at
\$\AstraCostPerK\ for Astra alone. That is \CostRatioAstraOverCascade$\times$ lower cost at
an accuracy this sample cannot resolve either way, so the claim is cost, its accuracy leg
being a failure to detect a difference rather than a demonstration that none exists. The
claim is also cost and not routing: escalating the same share at random reaches the same
place for \$\RandomAtNinetyCost\ (Section~\ref{subsec:random-control}), so the probability
has to earn its keep on the guard metric, which is where the rest of this section goes.

Matched accuracy is not matched behaviour: an operator choosing a guard buys the rate at
which unsupported claims get through, not the aggregate
(Section~\ref{sec:metrics}), and on that rate ours is the more permissive. Cross-fitted alongside its accuracy it false-verifies
\CascadeCrossfitFVRTwoDP\% of gold-unsupported claims against \AstraFVRTwoDP\%, a gap of
\CascadeAstraFVRDeltaPP\ points clearing both instruments of Section~\ref{sec:setup}
uncorrected but not the Holm correction the same section requires
($p = \CascadeAstraFVRMcNemarPHolm$, Appendix~\ref{app:limitations}), so we
report the direction, not a demonstrated separation
(Table~\ref{tab:cascade-vs-astra}). It inherits
Jev's verdict wherever it does not escalate, so the gate moves it partway from Jev's guard
behaviour towards Astra's: a property of the point an accuracy-maximising objective selects,
where one pricing the guard would take the tightest row of Table~\ref{tab:random}.

\subsection{The mechanism}

\begin{figure}[t]
\centering
\includegraphics[width=0.62\linewidth]{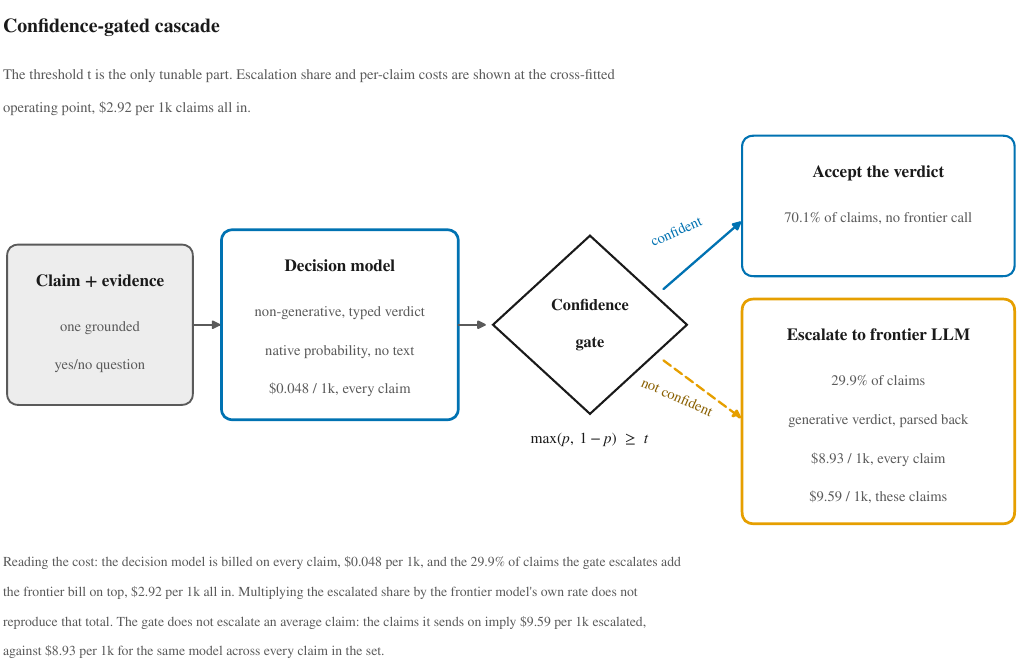}
\caption{Confidence-gated cascade. The accept and escalate shares shown are means over
the \CrossfitRepeats\ cross-fitting repeats rather than the shares produced by any single
fixed threshold, so they coincide with no one row of Table~\ref{tab:cascade}.}
\label{fig:architecture}
\end{figure}

Figure~\ref{fig:architecture} shows the arrangement. Each claim and its evidence go to
Jev, which returns a typed verdict with a native probability. When the confidence
$\max(p,\,1-p)$ clears the threshold, Jev's verdict is the pipeline's; otherwise the same
claim and evidence go to Astra. The cascade is simulated offline: every routing decision
recomposes locked predictions, and cost per thousand claims sums the per-claim costs along
each claim's route. Orchestration, retries and
the second network hop are not modelled, so the cost is a floor, not a quote.

\subsection{Choosing the threshold without looking at the answer}

We run \CrossfitRepeats\ repeats of stratified \CrossfitFolds-fold cross-fitting; within
each repeat the threshold maximising macro-balanced accuracy on the remaining folds is
applied to the held-out fold, so every claim receives exactly one out-of-fold routing
decision. Scoring is over the assembled \NPaired\ claims rather than per fold, because some
datasets contribute too few unsupported claims for a per-fold macro. Table~\ref{tab:cascade}
places the result against the single-system configurations.

\begin{table}[t]
\centering\small
\caption{Cascade against single systems. The cross-fitted row is the reportable operating
point; its bracket is the central 95\% percentile interval over the \CrossfitRepeats\
repeats and carries fold-assignment and threshold-selection noise only, so it is not a
confidence interval on a population difference. That row's false-verification rate comes
out of the same cross-fitting loop as its accuracy, so the two columns describe one
operating point, and it is the column on which the cascade loses to the frontier arm.
Oracle routing escalates exactly the claims Jev gets wrong and is an upper bound no
router can reach. The whole accuracy column is shown to two decimals: the high-effort Astra
row and the cascade row round to the same value at one while being different quantities,
and Section~\ref{sec:cascade-inference} differences two of these cells, which a reader is
entitled to reproduce from the column rather than from a rounding of it.}
\label{tab:cascade}
\begin{tabular}{lrrrr}
\toprule
Configuration & Macro BAcc & FVR (\%) & Escalated & \$/1k claims \\
\midrule
Jev alone                    & \JevBAccTwoDP             & \JevFVRTwoDP              & none                       & \JevCostPerK \\
Cascade, cross-fitted        & \CascadeCrossfitBAccTwoDP & \CascadeCrossfitFVRTwoDP  & \CascadeCrossfitEscalated\% & \CascadeCrossfitCost \\
Astra, standard effort       & \AstraBAccTwoDP           & \AstraFVRTwoDP            & all                        & \AstraCostPerK \\
Astra, high effort           & \AstraHighBAccTwoDP       & \AstraHighFVRTwoDP        & all                        & \AstraHighCostPerK \\
\midrule
Oracle routing (upper bound) & \OracleBAccTwoDP          & \OracleFVRTwoDP           & \OracleEscalated\%          & \OracleCost \\
\bottomrule
\end{tabular}
\end{table}

\subsection{Testing the comparison the cost claim rests on}
\label{sec:cascade-inference}

Saying two systems score the same needs the machinery that saying they differ needs, and
the bracket in Table~\ref{tab:cascade} is not it: it resamples the fold assignment, never
the claims. So the cascade goes through both instruments of Section~\ref{sec:setup}: the
bootstrap cycles through the assembled out-of-fold prediction vectors, carrying
fold-assignment noise rather than conditioning it away, and the exact McNemar runs once per
repeat, there being \CrossfitRepeats\ such vectors and no canonical one.

The cross-fitted cascade is \CascadeAstraDeltaPP\ points above Astra on macro-balanced
accuracy, 95\% CI [${\CascadeAstraDeltaCILo}$, \CascadeAstraDeltaCIHi]: an interval containing
zero and \CascadeAstraDeltaCIWidthPP\ points wide, wider than any difference that would
matter here, so \NPaired\ claims cannot resolve this. The test agrees, at a median exact
McNemar $p$ of \CascadeAstraMcNemarPMedian\ over repeats on a median of
\CascadeAstraDiscordantMedian\ discordant items. Nor is the sign stable: the cascade lands \emph{below} Astra on
macro-balanced accuracy in \CascadeAstraRepeatsMacroBelowAstra\ of the
\CrossfitRepeats\ repeats, and Appendix~\ref{app:limitations} has the rest of the counts.

The cascade does worst against the high-effort frontier arm, the neighbour beneath it in
Table~\ref{tab:cascade}. On accuracy that difference is unresolved again
(Table~\ref{tab:cascade-vs-astra}). It is
\CascadeAstraHighFVRDeltaPP\ points worse on the guard metric, 95\% CI
[\CascadeAstraHighFVRDeltaCILo, \CascadeAstraHighFVRDeltaCIHi], median $p$
\CascadeAstraHighFVRMcNemarPMedian\ and $\CascadeAstraHighFVRMcNemarPHolm$ adjusted over
the \MultiplicityFamilyN-test family, which makes it one of the
\MultiplicitySurvivingN\ corrected survivors and the only one about the cascade. That
guard costs \$\AstraHighCostPerK\ per thousand claims against \$\CascadeCrossfitCost.

\begin{table}[t]
\centering\small
\caption{The brackets the cascade carries, which answer different questions. The first row
is the spread of the cross-fitted estimate over fold draws and threshold selection on this
fixed sample. The rest are sampling uncertainty on the difference against each neighbour,
and only those license a statement about whether the systems differ. On accuracy none of
them does, since all three straddle zero. On the guard metric the comparison against Astra clears
both instruments of Section~\ref{sec:setup} uncorrected and sits on the $0.05$ boundary
once corrected ($p = \CascadeAstraFVRMcNemarPHolm$ Holm-adjusted over the
\MultiplicityFamilyN-test family), the comparison against Jev clears the interval
but not the test, and the comparison against the high-effort arm clears both instruments
and the correction ($p = \CascadeAstraHighFVRMcNemarPHolm$ adjusted). All of them come out
of the same cross-fitting loop over the same locked predictions.}
\label{tab:cascade-vs-astra}
\small
\begin{tabular}{llr}
\toprule
Quantity & What it varies & Value \\
\midrule
Cross-fitted macro BAcc, spread over repeats & fold draw, threshold choice
  & \CascadeCrossfitBAccTwoDP\ [\CascadeCrossfitCILo, \CascadeCrossfitCIHi] \\
Cascade minus Astra, 95\% CI & claims, fold draw
  & \CascadeAstraDeltaPP\ [${\CascadeAstraDeltaCILo}$, \CascadeAstraDeltaCIHi] \\
Exact McNemar $p$, median over repeats & fold draw
  & \CascadeAstraMcNemarPMedian\ (\CascadeAstraMcNemarPMin\ to \CascadeAstraMcNemarPMax) \\
Cascade minus Jev, 95\% CI & claims, fold draw
  & \CascadeJevDeltaPP\ [${\CascadeJevDeltaCILo}$, \CascadeJevDeltaCIHi] \\
Cascade minus high-effort Astra, 95\% CI & claims, fold draw
  & ${\CascadeAstraHighDeltaPP}$\ [${\CascadeAstraHighDeltaCILo}$, \CascadeAstraHighDeltaCIHi] \\
\quad exact McNemar $p$, median over repeats & fold draw
  & \CascadeAstraHighMcNemarPMedian\ (\CascadeAstraHighMcNemarPMin\ to \CascadeAstraHighMcNemarPMax) \\
\midrule
\emph{The guard metric, on the gold-unsupported claims} & \emph{same bootstrap, same test} & \\
Cascade minus Astra FVR, 95\% CI & claims, fold draw
  & \CascadeAstraFVRDeltaPP\ [\CascadeAstraFVRDeltaCILo, \CascadeAstraFVRDeltaCIHi] \\
\quad exact McNemar $p$, median over repeats & fold draw
  & \CascadeAstraFVRMcNemarPMedian\ (\CascadeAstraFVRRepeatsPBelowFive/\CrossfitRepeats\ below 0.05) \\
Cascade minus Jev FVR, 95\% CI & claims, fold draw
  & ${\CascadeJevFVRDeltaPP}$\ [${\CascadeJevFVRDeltaCILo}$, ${\CascadeJevFVRDeltaCIHi}$] \\
\quad exact McNemar $p$, median over repeats & fold draw
  & \CascadeJevFVRMcNemarPMedian\ (\CascadeJevFVRRepeatsPBelowFive/\CrossfitRepeats\ below 0.05) \\
Cascade minus high-effort Astra FVR, 95\% CI & claims, fold draw
  & \CascadeAstraHighFVRDeltaPP\ [\CascadeAstraHighFVRDeltaCILo, \CascadeAstraHighFVRDeltaCIHi] \\
\quad exact McNemar $p$, median over repeats & fold draw
  & \CascadeAstraHighFVRMcNemarPMedian\ (\CascadeAstraHighFVRRepeatsPBelowFive/\CrossfitRepeats\ below 0.05) \\
\bottomrule
\end{tabular}
\end{table}

\subsection{The cascade against the cheap system it is made of}
\label{sec:cascade-vs-jev}

Everything above compares the cascade with the expensive arm, the comparison that flatters
it; the other has to be printed too. It costs \CostRatioCascadeOverJev$\times$ what Jev
alone costs, since every claim still pays Jev and \CascadeCrossfitEscalated\% then pay the
frontier price on top.

On accuracy this sample does not make the case. The cross-fitted cascade is
\CascadeCrossfitMinusJevPP\ points above Jev alone, 95\% CI [${\CascadeJevDeltaCILo}$,
\CascadeJevDeltaCIHi], at a median exact McNemar $p$ of \CascadeJevMcNemarPMedian\ over
the \CrossfitRepeats\ repeats, and it lands below Jev in
\CascadeJevRepeatsMacroBelowJev\ of them. The point estimate is the size of the run-to-run
movement between two passes of Jev over identical claims
(Section~\ref{sec:limitations}), the scale at which a \CostRatioCascadeOverJev-fold cost
increase would have to be justified and is not.

What it buys over Jev is the guard metric, the only positive claim
this paper makes for the arrangement over its own cheap tier, so it gets the machinery
rather than an assertion. Cross-fitted, it false-verifies fewer gold-unsupported claims than
Jev alone, \CascadeCrossfitFVRTwoDP\% against \JevFVRTwoDP\%, a change of
${\CascadeJevFVRDeltaPP}$\ points whose 95\% interval [${\CascadeJevFVRDeltaCILo}$,
${\CascadeJevFVRDeltaCIHi}$] excludes zero. The test does not follow: the exact McNemar on
the discordant false verifications has a median $p$ of \CascadeJevFVRMcNemarPMedian\ and
falls below $0.05$ in only \CascadeJevFVRRepeatsPBelowFive\ of the repeats. One instrument
clears and the other does not, so by the criterion of Section~\ref{sec:setup} this is a
consistent improvement of marginal significance, not a demonstrated benefit. The direction
never reverses in all \CascadeJevFVRRepeatsFavouringCascade\ repeats, the claims it
escalates being the ones Jev is least sure about. It sits between the two systems on both axes: part of
Astra's guard for part of Astra's price.

\subsection{Against spending the same budget at random}
\label{subsec:random-control}

\begin{figure}[t]
\centering
\includegraphics[width=0.62\linewidth]{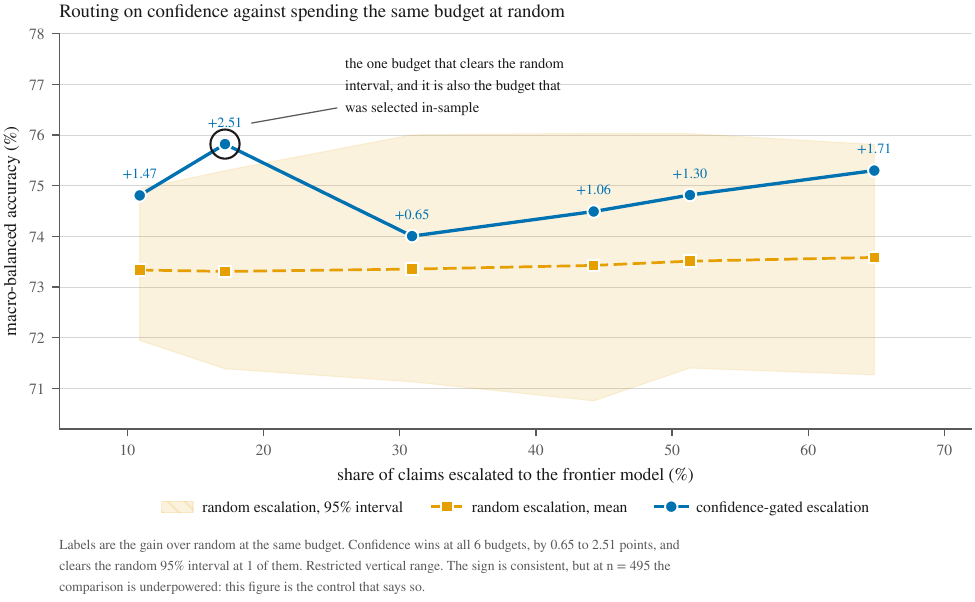}
\caption{Confidence routing against the matched random-escalation control, with the
random baseline's interval shaded. The vertical axis is restricted rather than running
from zero to one hundred, so the visual separation between the two curves is smaller than
it appears.}
\label{fig:routing}
\end{figure}

Without a control, an apparent cascade gain could be nothing but sending extra
traffic to the stronger system, so for each escalation share we also route that share
uniformly at random (Table~\ref{tab:random}, Figure~\ref{fig:routing}). Confidence routing
beat it at every budget tested, by \ConfMinusRandomMinPP\ to \ConfMinusRandomMaxPP\ points
(\NRandomBudgetsConfidenceWins\ of \NRandomBudgetsTested), and the power is not there:
only the \RandomAtEightyEscalated\% budget clears the random interval, and that is the one
the full sample selects, so it is the row most exposed to selection. We report
\NRandomBudgetsConfidenceWins\ consistent signs and nothing stronger.

That control also prices the cost claim. At the budget nearest the reported operating
point, escalating \RandomAtNinetyEscalated\% of claims at random costs
\$\RandomAtNinetyCost\ per thousand against \$\CascadeAtNinetyCost\ for the confidence
gate, scoring \RandomAtNinetyBAccTwoDP\ against the frontier
arm's \AstraBAccTwoDP\ at \$\AstraCostPerK. Overlapping two levels would prove nothing,
so the control goes through the criterion of Section~\ref{sec:setup} under both its
instruments, as every other comparison here does: the paired difference is
$\RandomVsAstraDeltaPP$ points, [$\RandomVsAstraDeltaCILo$, $\RandomVsAstraDeltaCIHi$],
cycling the random draws so routing noise is carried, at exact McNemar
$p = \RandomVsAstraMcNemarPMedian$ on the median draw,
\RandomVsAstraRepeatsPBelowFive\ of \CrossfitRepeats\ draws below 0.05. A coin flip at
that budget is not separable from the frontier model either, so the saving is a property of
the escalation share, not of the probability that picks what to escalate. What that
probability buys is the \ConfMinusRandomMinPP\ to \ConfMinusRandomMaxPP\ points above,
which this sample cannot separate from zero.

\begin{table}[t]
\centering\small
\caption{Confidence routing against random escalation at a matched budget, with the guard
metric and the bill for each operating point. Random intervals are central 95\% percentile
intervals over random draws at the same escalation share. Every row of the confidence
column uses a threshold fitted on the full sample, so all of it carries the selection
optimism quantified in Section~\ref{sec:tuned} and none of it is an out-of-sample
operating point; the cross-fitted point of Table~\ref{tab:cascade}, which escalates
\CascadeCrossfitEscalated\% of claims, appears on no row here. The FVR and cost columns
are printed because the accuracy column alone would hide what
Section~\ref{sec:discussion} tells operators to buy on, and because the tightest row is
the one that matters for that reading: at \RandomAtNinetyNineEscalated\% escalation the
cascade false-verifies \CascadeAtNinetyNineFVRCount\ of the \NGoldUnsupported\
gold-unsupported claims against \AstraFVRCount\ for the frontier arm, a margin of a
single claim on which the exact McNemar returns
$p = \CascadeAtNinetyNineFVRVsAstraMcNemarP$, while scoring
\CascadeAtNinetyNineBAcc\ against \AstraBAccTwoDP\ and costing \$\CascadeAtNinetyNineCost\
against \$\AstraCostPerK. Its accuracy margin over Astra,
\CascadeAtNinetyNineMinusAstraPP\ points, sits inside the
\CascadeAstraDeltaCIWidthPP-point interval this paper reports for that estimand, and the
high-effort frontier configuration false-verifies \AstraHighFVRCount\ of the same claims,
fewer than this row does. So the tightest row matches the frontier arm's guard to within a
single claim while scoring higher in sample and costing less, and cost is the only one of the
three axes this sample resolves there. A one-claim margin decides nothing either way. Our
procedure did not select this row: it maximises macro-balanced accuracy and lands on
\CascadeInSampleThreshold, so Section~\ref{sec:tuned} applies to any operator who picks this
row for its guard as much as it applies to ours.}
\label{tab:random}
\small
\begin{tabular}{rrrrr}
\toprule
Escalated (\%) & Confidence & Random [95\% interval] & FVR (\%) & \$/1k \\
\midrule
\RandomAtSeventyEscalated      & \CascadeAtSeventyBAccTwoDP      & \RandomAtSeventyBAccTwoDP\ [\RandomAtSeventyCILoTwoDP, \RandomAtSeventyCIHiTwoDP]           & \CascadeAtSeventyFVR      & \CascadeAtSeventyCost \\
\RandomAtEightyEscalated       & \CascadeAtEightyBAccTwoDP       & \RandomAtEightyBAccTwoDP\ [\RandomAtEightyCILoTwoDP, \RandomAtEightyCIHiTwoDP]              & \CascadeAtEightyFVR       & \CascadeAtEightyCost \\
\RandomAtNinetyEscalated       & \CascadeAtNinetyBAccTwoDP       & \RandomAtNinetyBAccTwoDP\ [\RandomAtNinetyCILoTwoDP, \RandomAtNinetyCIHiTwoDP]              & \CascadeAtNinetyFVR       & \CascadeAtNinetyCost \\
\RandomAtNinetyFiveEscalated   & \CascadeAtNinetyFiveBAccTwoDP   & \RandomAtNinetyFiveBAccTwoDP\ [\RandomAtNinetyFiveCILoTwoDP, \RandomAtNinetyFiveCIHiTwoDP]  & \CascadeAtNinetyFiveFVR   & \CascadeAtNinetyFiveCost \\
\RandomAtNinetySevenEscalated  & \CascadeAtNinetySevenBAccTwoDP  & \RandomAtNinetySevenBAccTwoDP\ [\RandomAtNinetySevenCILoTwoDP, \RandomAtNinetySevenCIHiTwoDP] & \CascadeAtNinetySevenFVR  & \CascadeAtNinetySevenCost \\
\RandomAtNinetyNineEscalated   & \CascadeAtNinetyNineBAccTwoDP   & \RandomAtNinetyNineBAccTwoDP\ [\RandomAtNinetyNineCILoTwoDP, \RandomAtNinetyNineCIHiTwoDP]  & \CascadeAtNinetyNineFVR   & \CascadeAtNinetyNineCost \\
\bottomrule
\end{tabular}
\end{table}

\subsection{How much room the idea has}

\begin{figure}[t]
\centering
\includegraphics[width=0.62\linewidth]{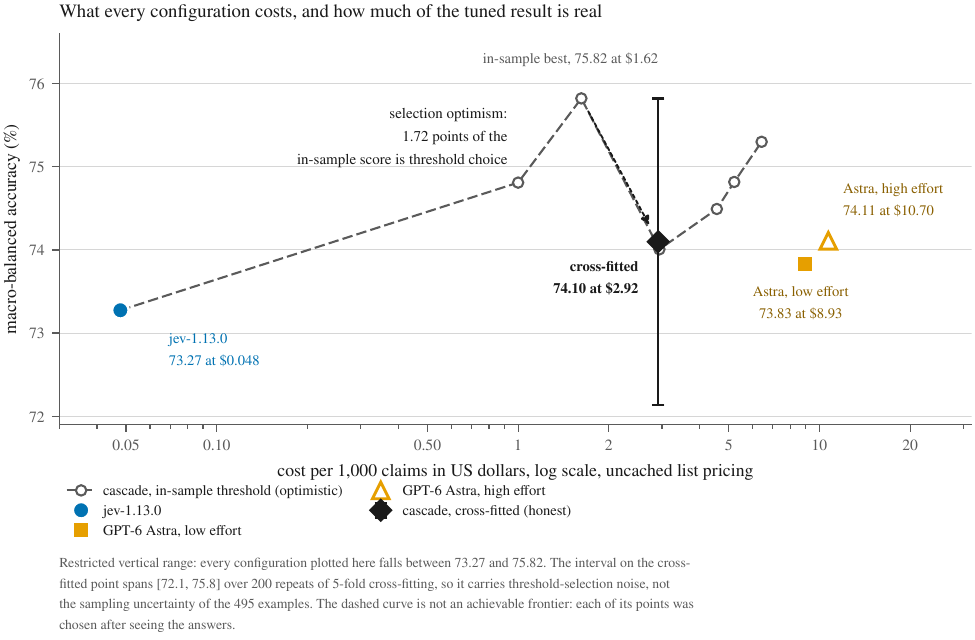}
\caption{Cost-accuracy frontier. Cascade operating points are drawn as a curve and the
single-system configurations as points, with the in-sample and cross-fitted cascade
plotted distinctly so the selection optimism of Section~\ref{sec:tuned} is visible. The
vertical axis is restricted rather than running from zero to one hundred.}
\label{fig:frontier}
\end{figure}

Oracle routing, which escalates exactly the claims Jev gets wrong, scores \OracleBAcc\ by
escalating \OracleEscalated\% of claims at \$\OracleCost\ per thousand,
\OracleHeadroomPP\ points above Jev alone. It needs the answer first, so it bounds the idea
rather than offering an operating point. A perfect ranking of Jev's own errors would
collect all that headroom; \JevSelectiveAUROC\ collects a slice, enough for an accuracy
this sample cannot resolve against Astra's and not enough to put a detectable gain inside
the interval above. Ranking its own errors as well as it ranks claims would move the
cascade toward the oracle line of Figure~\ref{fig:frontier} with neither system changing.

\section{How much of a tuned cascade is real}
\label{sec:tuned}

Choosing the escalation threshold on the same \NPaired\ claims the cascade is scored
on returns \CascadeInSampleBAccTwoDP\ macro-balanced accuracy, against \CascadeCrossfitBAccTwoDP\
when the threshold is chosen out of fold. The difference, \SelectionOptimismPP\ points, is
selection and nothing else: same predictions, same sweep, same metric, one fitted threshold.
The in-sample point flatters the cost axis too, escalating \CascadeInSampleEscalated\% of
claims at \$\CascadeInSampleCost\ per thousand against the cross-fitted
\CascadeCrossfitEscalated\% at \$\CascadeCrossfitCost, so a tuned result promises an
operating point more accurate and cheaper than the procedure reproduces on new data.

\subsection{The threshold is not identifiable at this sample size}

Across \CrossfitThresholdFits\ fold-fits the procedure chose
\CascadeInSampleThreshold\ \ThresholdPickedEighty\ times and the grid's highest
threshold \ThresholdPickedNinetyNine, taking the five remaining points far less
often: \ThresholdPickedFifty, \ThresholdPickedSeventy, \ThresholdPickedNinety,
\ThresholdPickedNinetyFive\ and \ThresholdPickedNinetySeven\ times, in grid order. The
two winners are different regimes, not neighbouring settings, escalating
\CascadeAtEightyEscalated\% of claims and \CascadeAtNinetyNineEscalated\%. The mean
escalation rate of \CascadeCrossfitEscalated\% in Table~\ref{tab:cascade} averages over that
bimodal choice, so it is the procedure's expected spend rather than a threshold anyone
would deploy. At \NPaired\ claims the optimum is a coin flip between two regimes.

The bracket on the cross-fitted estimate, [\CascadeCrossfitCILo, \CascadeCrossfitCIHi],
carries fold assignment and threshold selection but not the sampling uncertainty of the
claims, so it is a lower bound on that number's total uncertainty. Its upper end needs
explaining: \CascadeCrossfitCIHi\ is the in-sample best this section calls an
artefact, printed above as \CascadeInSampleBAccTwoDP\ and here at the bracket's one
decimal. A repeat that picks
\CascadeInSampleThreshold\ in all \CrossfitFolds\ folds reconstructs the in-sample cascade
exactly, and at the recorded seed that happens in \RepeatsReproducingInSample\
repeats, \RepeatsReproducingInSamplePct\%, more than the $2.5\%$ a $97.5$th percentile needs.
Nothing pins that share there and we measured no other seeds, so it is a property of this
draw of folds rather than something to explain away.

Every cross-fitted mean here is one draw of the fold assignment at one recorded seed, with a
Monte Carlo error that Appendix~\ref{app:limitations} states per quantity.

\subsection{What this costs, and where the number belongs}

FrugalGPT, Hybrid LLM and RouteLLM each fit their routing rule on data they do not then
score on \citep{chen2023frugalgpt, ding2024hybridllm, ong2024routellm}; that discipline is
what Section~\ref{sec:cascade}'s cross-fitting applies to a threshold. Where fitting and
scoring do share data the reported accuracy carries selection optimism, a hazard documented
for two decades \citep{varma2006bias, cawley2010selectionbias}, and the instance we can name
is our own companion article's in-sample sweep (Appendix~\ref{app:incidents}). On our sample
that inflates accuracy by \SelectionOptimismPP\ points and quotes \$\CascadeInSampleCost\ per
thousand claims where the reproducible figure is \$\CascadeCrossfitCost. How large it is
depends on the sweep, the sample and the signal, so ours is a measurement, not a constant;
cross-fitting prices the tuning, and that price belongs beside the
result. The honest
version of our own headline is the cross-fitted row of Table~\ref{tab:cascade}: a cost
reduction of \CostRatioAstraOverCascade$\times$, bought at
\CascadeAstraFVRDeltaPP\ points more false verification, with an accuracy difference against
the frontier arm this sample places anywhere in [${\CascadeAstraDeltaCILo}$,
\CascadeAstraDeltaCIHi].

%% file: close.tex
%
%

\section{Discussion}
\label{sec:discussion}

Everything above was measured on one benchmark with one pair of systems, and what it changes
for an engineer needs no retraining.

\subsection{Measure the error you actually pay for}

Two systems within \JevAstraDeltaPP\ points of each other on macro-balanced accuracy differ
by \FVRGapPP\ points on the rate that decides whether a fabricated sentence reaches a user:
Jev verifies \JevFVR\% of the unsupported claims against \AstraFVR\% for Astra. A product that puts a green check beside a verified answer pays
for a false verification in user trust and for a false alarm in review-queue minutes; those
prices are rarely equal, so an aggregate weighting them equally is a poor objective.
Report the false-verification rate and verified precision at the operating point you
intend to ship, alongside whatever aggregate the literature expects; on the headline
alone these two systems look interchangeable.

\subsection{Route rather than replace, only against the frontier arm}

Because the two systems fail on different claims (Section~\ref{sec:different-errors}),
the cheap one answers what it is confident about and escalates the rest. Which arm it
beats is the recommendation: over the frontier model routing pays, over Jev alone
it is not justified on this sample. Cross-fitted it reaches \CascadeCrossfitBAccTwoDP\
macro-balanced accuracy at \$\CascadeCrossfitCost\ per thousand claims against
\AstraBAccTwoDP\ at \$\AstraCostPerK, a comparison this sample cannot resolve
(Section~\ref{sec:cascade-inference}); against the tier it is built from it costs
\CostRatioCascadeOverJev-fold as much for \CascadeCrossfitMinusJevPP\ points, 95\% CI
[${\CascadeJevDeltaCILo}$, \CascadeJevDeltaCIHi] (Section~\ref{sec:cascade-vs-jev}). What
routing resolves is the bill, and not even that is the signal's doing: spending the same
budget at random costs \$\RandomAtNinetyCost\ and is no further from the frontier arm
(Section~\ref{subsec:random-control}), so what to carry away is the escalation share. On the
guard the signal points one way without settling it. The cascade false-verifies \CascadeAstraFVRDeltaPP\ points more
often than the frontier arm, 95\% CI [\CascadeAstraFVRDeltaCILo,
\CascadeAstraFVRDeltaCIHi], clearing both instruments uncorrected but not the Holm
correction ($p = \CascadeAstraFVRMcNemarPHolm$, Appendix~\ref{app:limitations}). So
it buys a \CostRatioAstraOverCascade-fold cost reduction against that arm and
probably pays for it on the axis an operator least wants to pay on. A team needing
the tighter guard should read Table~\ref{tab:gating}, or the tightest row of
Table~\ref{tab:random}, where the cascade matches the frontier arm's false verifications to
within a claim at \$\CascadeAtNinetyNineCost\ per thousand: selected in sample,
deciding nothing either way, extending neither result (Section~\ref{sec:tuned}).

\subsection{Where else a decision model fits}

A pipeline that asks a yes/no question at volume and pays a text generator to answer in
prose is paying for the prose. Grounded verification is one such question; routing,
moderation, retrieval triage and schema validation share the shape, and the arithmetic
behind the \CostRatioAstraOverJev-fold does not depend on the
question being about entailment. We measured one question on one benchmark, so the transfer
is a hypothesis with a cheap test attached, not a result.

\section{Limitations and Conflict of Interest}
\label{sec:limitations}

Every number here comes from \NPaired\ paired claims on a public benchmark's development
split, every prompt, threshold, grid and metric chosen with these outcomes visible; the
test split was never read. Appendix~\ref{app:limitations} carries each item's numbers,
in this order.

No equivalence is claimed or tested. \NClearingUncorrected\ comparisons meet the separation criterion of
Section~\ref{sec:setup} uncorrected: the false-verification gap between the single systems
and the cascade's guard penalty against each frontier arm; \MultiplicitySurvivingN\ survive
Holm-Bonferroni over the \MultiplicityFamilyN\ tests here, and no other among the reported
configurations meets it either way. Both tests against the excluded MiniCheck arm clear the
adjustment too, as evidence for that exclusion. One threshold is fitted out of fold and nothing
else is: the gating sweep and tuned cascade are in-sample by construction, grid and
objective fixed on these data. The cross-fitted cascade's sign is unstable too, below Astra
on accuracy in \CascadeAstraRepeatsMacroBelowAstra\ of \CrossfitRepeats\ repeats. The
evaluation sample is a released list of example IDs no code regenerates, from an exploratory
Jev run whose retained rows all succeeded, so its failure rate is unmeasurable here. The cascade is an offline replay, not a deployed router; its cost excludes the second hop,
retries and queueing. Costs are measured tokens at archived list prices, not invoices; neither
vendor's volume rates are public, and free output tokens are policy, not
architecture. Each arm ran once; the frontier repeat covers only
rows reporting no internal reasoning, and an earlier Jev pass flipped \JevRerunFlips\ verdicts,
\JevRerunDeltaPP\ points lower. No latency figure appears, the bulk runs being concurrent.
Contamination cannot be ruled out, and the frontier side is one model family at two efforts
under one instruction forbidding reasoning, with no provider-default run. A third arm was excluded: our MiniCheck port failed
as our harness anchor, so we do not call the harness sound. The calibration comparison has one participant, Astra having none at all. Nothing was truncated and no example dropped for length, so it says
nothing about either context limit. The harness was written with AI coding assistance; the metric layer is independently
reimplemented, the pipeline above it not.

\subsection{Conflict of interest}

The first author founded and leads CustomGPT.ai, whose
product includes claim verification \citep{customgpt2026verify, customgpt2026results,
customgpt2026verifyapi}; the second author works there. CustomGPT.ai is a paying API customer of both vendors, with no
discount, credit, early access, rate-limit accommodation or engineering support; neither was
contacted, had input into the design, data, analysis or results, or reviewed this paper. For
TypeSafe the released package carries the clause check, the sha256 of the capture it was read
from, the unsent permission request and the decision that followed. The agreement is the
vendor's and is not republished, so what a reader checks is the live clause rather than our
hash (Appendix~\ref{app:limitations}, with the order and the dates). We found no clause
restricting publication of benchmark results. For the frontier vendor, the authors' attestation. Our commercial interest in cheap verification is why every prediction, prompt and
line of analysis is released (Appendix~\ref{app:repro}).

\section{Conclusion}
\label{sec:conclusion}

A non-generative decision model and a frontier LLM landed within a point on grounded claim
verification, \JevBAccTwoDP\ against \AstraBAccTwoDP\ macro-balanced accuracy on
\NPaired\ paired claims, Jev minus Astra ${\JevAstraDeltaSignedTwoDP}$ points, not
separable from zero, at \CostRatioAstraOverJev-fold the cost. The matching
headline concealed the useful fact: they get different claims wrong. They agree on
\AgreementPct\% of decisions, Jev is right where Astra errs on \JevOnlyCorrect\ claims
and the reverse on \AstraOnlyCorrect, and their false-verification rates differ by
\FVRGapPP\ points. Errors that differ compose, and escalating the least confident
\CascadeCrossfitEscalated\% of claims cut the bill \CostRatioAstraOverCascade-fold,
\$\CascadeCrossfitCost\ per thousand claims against \$\AstraCostPerK, at
\CascadeAstraFVRDeltaPP\ points more false verification, 95\% CI
[\CascadeAstraFVRDeltaCILo, \CascadeAstraFVRDeltaCIHi], at an accuracy this sample cannot
resolve against the frontier model's. That saving belongs to the escalation budget, not to
the signal spending it: the nearest random-escalation budget costs \$\RandomAtNinetyCost.
The extra false verification is its price, not a second benefit. Against Jev alone it costs
\CostRatioCascadeOverJev-fold as much for \CascadeCrossfitMinusJevPP\ points, 95\% CI
[${\CascadeJevDeltaCILo}$, \CascadeJevDeltaCIHi], so it replaces the frontier model, not
the cheap tier. Evaluated as tuned cascades usually are it returns
\CascadeInSampleBAccTwoDP, \SelectionOptimismPP\ points of selection rather than
capability. Oracle routing reaches \OracleBAcc, where the next result here is hiding.

\appendix

\section{Per-dataset results}
\label{app:per-dataset}

Table~\ref{tab:per-dataset} gives balanced accuracy per component dataset for
the three scored configurations, with the gold class support behind each row. The
minority class is thin in several datasets, so these rows describe where the
aggregate comes from and do not support per-dataset inference.

Section~\ref{sec:different-errors} says that the per-dataset differences between Jev and
Astra cancel almost entirely, and here is the arithmetic. Every difference below is computed
from the unrounded accuracies rather than from the rounded cells of
Table~\ref{tab:per-dataset}, so subtracting two of those cells can land in the last digit
rather than on the number printed here. The two widest cells, AggreFact-CNN
and ExpertQA, are ${\DeltaAggreFactCNNPP}$\ and \DeltaExpertQAPP\ points: their magnitudes sum
to \WidestTwoSumAbsPP\ points against a signed ${\WidestTwoSumPP}$, so they cancel
\WidestTwoCancellationPct\% of themselves. The other \NRemainingDatasets\ are not small
either. Their magnitudes sum to \RemainingSumAbsPP\ points, the largest is
\MaxAbsRemainingPP\ and the mean \MeanAbsRemainingPP, and they cancel
\RemainingCancellationPct\% of themselves, down to a signed ${\RemainingNineSumPP}$. The
per-dataset disagreement that Section~\ref{sec:different-errors} quotes is these two
magnitude sums added before either was rounded. All three figures are rounded to one decimal
independently of each other, so adding the two printed parts need not land on the printed
total, and where it does not, the total is the sum and the parts are the rounding. The further arithmetic that
\WidestTwoShareOfSumPct\% of the signed total comes from two cells is a property of summing
signed numbers that nearly annihilate and carries no weight of its own, which is why
Section~\ref{sec:different-errors} argues from the magnitudes instead.

\begin{table}[htbp]
\centering
\small
\caption{Balanced accuracy by component dataset, with gold class support. Sup.\ and
Uns.\ are the gold-supported and gold-unsupported counts in the sampled claims.}
\label{tab:per-dataset}
\begin{tabular}{lrrrrrr}
\toprule
Dataset & $n$ & Sup. & Uns. & Jev & Astra (low) & Astra (high) \\
\midrule
AggreFact-CNN    & \SupportAggreFactCNNN    & \SupportAggreFactCNNPos    & \SupportAggreFactCNNNeg    & \JevBAccAggreFactCNN    & \AstraBAccAggreFactCNN    & \AstraHighBAccAggreFactCNN    \\
AggreFact-XSum   & \SupportAggreFactXSumN   & \SupportAggreFactXSumPos   & \SupportAggreFactXSumNeg   & \JevBAccAggreFactXSum   & \AstraBAccAggreFactXSum   & \AstraHighBAccAggreFactXSum   \\
ClaimVerify      & \SupportClaimVerifyN     & \SupportClaimVerifyPos     & \SupportClaimVerifyNeg     & \JevBAccClaimVerify     & \AstraBAccClaimVerify     & \AstraHighBAccClaimVerify     \\
ExpertQA         & \SupportExpertQAN        & \SupportExpertQAPos        & \SupportExpertQANeg        & \JevBAccExpertQA        & \AstraBAccExpertQA        & \AstraHighBAccExpertQA        \\
FactCheck-GPT    & \SupportFactCheckGPTN    & \SupportFactCheckGPTPos    & \SupportFactCheckGPTNeg    & \JevBAccFactCheckGPT    & \AstraBAccFactCheckGPT    & \AstraHighBAccFactCheckGPT    \\
LFQA             & \SupportLfqaN            & \SupportLfqaPos            & \SupportLfqaNeg            & \JevBAccLfqa            & \AstraBAccLfqa            & \AstraHighBAccLfqa            \\
RAGTruth         & \SupportRAGTruthN        & \SupportRAGTruthPos        & \SupportRAGTruthNeg        & \JevBAccRAGTruth        & \AstraBAccRAGTruth        & \AstraHighBAccRAGTruth        \\
Reveal           & \SupportRevealN          & \SupportRevealPos          & \SupportRevealNeg          & \JevBAccReveal          & \AstraBAccReveal          & \AstraHighBAccReveal          \\
TofuEval-MediaS  & \SupportTofuEvalMediaSN  & \SupportTofuEvalMediaSPos  & \SupportTofuEvalMediaSNeg  & \JevBAccTofuEvalMediaS  & \AstraBAccTofuEvalMediaS  & \AstraHighBAccTofuEvalMediaS  \\
TofuEval-MeetB   & \SupportTofuEvalMeetBN   & \SupportTofuEvalMeetBPos   & \SupportTofuEvalMeetBNeg   & \JevBAccTofuEvalMeetB   & \AstraBAccTofuEvalMeetB   & \AstraHighBAccTofuEvalMeetB   \\
Wice             & \SupportWiceN            & \SupportWicePos            & \SupportWiceNeg            & \JevBAccWice            & \AstraBAccWice            & \AstraHighBAccWice            \\
\bottomrule
\end{tabular}
\end{table}

\section{Prompts}
\label{app:prompts}

Both systems received the same criteria for SUPPORTED and NOT\_SUPPORTED, in the
native input format of each. The frontier prompt is a text template; the Jev prompt
is a typed question definition whose answer is an enumerated value, so there is no
output-format instruction to give it.

\subsection{Frontier prompt}

Used for GPT-6 Astra at both reasoning efforts, with the enum suffix appended. The
phrase \texttt{names the wrong entity} was added during the parity fix described in
Appendix~\ref{app:incidents}.

\begin{verbatim}
You are a grounded factuality classifier.

Decide whether the EVIDENCE supports every materially important part of the CLAIM.

SUPPORTED:
The evidence explicitly states or directly entails every materially important
part of the claim.

NOT_SUPPORTED:
Use this if any materially important part is missing, only partially supported,
contradicted, numerically wrong, temporally wrong, names the wrong entity, scoped
too broadly, or adds an unsupported causal/detail claim.

Judge only from the supplied evidence. Do not use outside knowledge.

CLAIM:
{claim}

EVIDENCE:
{evidence}

Respond with exactly one word, either SUPPORTED or NOT_SUPPORTED.
No explanation. No reasoning. No other text.
\end{verbatim}

\subsection{Jev question definition}

\begin{verbatim}
question_name: grounding
question_type: choice
instructions:
  Determine whether the evidence supports all materially important parts of
  the claim. Judge only from the supplied evidence.
criteria:
  supported:
    The evidence explicitly states or directly entails every materially
    important part of the claim.
  not_supported:
    The evidence fails to support at least one materially important part of
    the claim. This includes missing information, partial support,
    contradiction, incorrect numbers/dates/entities, scope overreach, or
    unsupported causal/detail additions.
state_keys: [claim, evidence]
label_map: {supported: SUPPORTED, not_supported: NOT_SUPPORTED}
\end{verbatim}

\section{Incidents and corrections}
\label{app:incidents}

What went wrong or was changed during this study is recorded here, because each of
these could otherwise be read off the released artifacts as a result.

\paragraph{A harness token cap, mistaken for model failures.}
An early paired run of GPT-6 Astra returned a large batch of empty responses and a
handful of provider errors. The cause was ours: the completion-token cap in our model
configuration was consumed by the model's internal reasoning tokens, leaving no budget
for the answer, and every failed call reported an output-token count exactly equal to
the cap while successful calls used a small fraction of it. The run was repeated with
a cap large enough to hold both. That failure rate is a property of our configuration
and appears nowhere as a model defect.

\paragraph{The prompt-parity fix.}
The Jev criteria listed incorrect entities as grounds for NOT\_SUPPORTED while the
frontier template listed only numeric and temporal errors, an asymmetry that favoured
Jev, since entity-swap errors are common in the AggreFact components. A
pre-publication audit of our own prompts caught it before any result was published.
We added \texttt{names the wrong entity} to the frontier template and repeated the
GPT-6 Astra runs at both reasoning efforts. Every Astra number in this paper comes
from the repeated runs, and what the edit moved on each arm, including the flip that
reached a gold-unsupported claim, is in Section~\ref{app:limitations}.

\paragraph{The MiniCheck exclusion.}
Described in Section~\ref{sec:limitations}. Its predictions are released with the
others so that a reader can diagnose the port rather than take our word for why it
was dropped.

\paragraph{The evidence file was edited after it was stamped.}
The hash file this paper's provenance claim rests on, and which the OpenTimestamps proof
covers, was rewritten while the analysis was being extended to hash two additional inputs.
That broke \texttt{ots verify} on the very artefact Appendix~\ref{app:repro} points a
reader at, and for a period this paper's own reproducibility claim would have failed if
checked. No result was affected: the prediction-file block was byte-identical across both
versions, so no number was regenerated. The stamped file was restored, the two additional
hashes moved to the separate, unstamped
\texttt{arxiv/evidence/input\_hashes.json}, and the writer now refuses to modify the
stamped file at all. It was caught by a verifier running \texttt{ots verify} rather than
reading our sentence about it.

\paragraph{The escalation rule in the code was not the rule in this paper.}
Section~\ref{sec:cascade} states the gate as $\max(p,\,1-p)$ reaching the threshold. The
analysis implemented the algebraically identical \texttt{(p >= t) or (p <= 1 - t)}, which
is not identical in binary floating point: \texttt{1 - 0.8} evaluates to
\texttt{0.19999999999999996}, so the one claim in this sample whose probability is exactly
\texttt{0.2} was escalated by the code and kept by the stated rule. A reader
reimplementing the sentence could not have reproduced our escalation rate. The code was
changed to follow the paper rather than the reverse, and every number was regenerated from
the unchanged locked predictions. The headline did not move. The escalation shares, their
costs and part of the gating sweep did, and the figures in this paper are the regenerated
ones; the incident log released with the package lists every value that changed, before and
after.

\paragraph{A clipping constant described as preregistered, and not.}
An earlier draft of Section~\ref{sec:calibration} called the log-loss clipping constant
preregistered. It is not: \texttt{configs/benchmark.yaml} marks it provisional, and no
preregistration fixes it. The sentence was retracted, the constant is now named as ours to
choose, and the section reports what the same predictions score at a clip matched to the
vendor's reported probability resolution instead of asserting one number.

\paragraph{The operating point we published first.}
A companion article by the first author was published before this paper, and it is in the
released package as \texttt{medium/article-v3.md} with its link in
\texttt{medium/story-link.md}. Its cascade figure is the in-sample threshold sweep, and the
row it leans on is the best one, which this paper recomputes as
\CascadeInSampleBAccTwoDP\ macro-balanced accuracy at
\CascadeInSampleEscalated\% escalation for \$\CascadeInSampleCost\ per thousand claims.
The article prints that row at one decimal and at the cost it had before the
escalation-rule correction above, which the released incident log records to the cent. That row is what
Section~\ref{sec:tuned} calls selection optimism, and it is superseded here by the
cross-fitted \CascadeCrossfitBAccTwoDP\ at \$\CascadeCrossfitCost, a gap of
\SelectionOptimismPP\ points of accuracy and most of the cost. The article states that the
sweep's accuracy column is not ordered and that picking a flattering row would be
objectionable, and then quotes that row's cost; the practice whose cost
Section~\ref{sec:tuned} measures is therefore one
this author has published, and it is the only instance that section names, because it is the
one we can document. That is why the correction is recorded here rather than left
for a reader to find by comparing the two documents. The article also predates the
escalation-rule fix above and carries the pre-fix figures for the \texttt{t=0.90} sweep
row. No number in this paper comes from it.

\paragraph{The released package was published stale, and the check that should have seen it
reported a status.} The model registry described above was corrected in our repository 77
minutes after the public reproduction package of Appendix~\ref{app:repro} was last built
and pushed. Until the package was rebuilt, a reader following that appendix to the pinned
registry read the completion-token cap of the discarded first pass, with none of the notes
the correction added. Three more files in that package were the pre-correction copies, one
of them \texttt{scripts/verify/} \texttt{check\_public\_release.sh} itself, which then
compared HTTP statuses rather than bytes and so answered PASS on the very registry it was
meant to catch, and a fourth, the registry's new guard test, was missing from the package
while that script's own list of paths named it. No result changes: the caps document runs
locked and hashed before the correction. What was wrong was the artefact our
reproducibility claim rests on, which is the reason this is recorded as a defect rather
than as housekeeping. The check now compares each published file by \texttt{sha256}, it
has to exit zero before this paper is submitted, and it is what found this.

These eight entries and the project's full incident log are the same set: the log ships in the
released package as \texttt{docs/reference/} \texttt{incident-log.md} and carries an entry
for each of them, with the before-and-after values this appendix summarises.

\section{Reproduction}
\label{app:repro}

Every number in this paper is regenerated by one command, which reads the locked
prediction files, the hashed configuration and the gold labels, and writes the macro file
that the paper is typeset against. The evaluation sample is an input to that command rather
than an output of it: the example-ID list is released and hashed with the predictions, and
the draw that produced it is not regenerated here (Section~\ref{sec:limitations}). The gold
labels are an input in the same sense, and they are the one input the released package does
not contain. The command fetches the pinned LLM-AggreFact dev parquet from the Hugging Face
dataset repository named in \texttt{data/manifests/} \texttt{aggrefact\_revision.json}, at
the revision recorded there. That dataset is gated: the command needs \texttt{HF\_TOKEN}
set to a token whose account has accepted the dataset's terms on its Hugging Face page, and
without one the first read fails with a 401 rather than producing wrong numbers. We do not
redistribute the labels because they are not ours to redistribute, and the file's sha256 is
in the stamped evidence lock below, so a reader can check that what they download is what
we scored against. The package's
\texttt{.env.example} names the token and the terms-acceptance step, and the same file
records that \texttt{docker-compose.yml} defaults the container user to uid 1000, which
has to be changed to the host's own uid on a machine where it differs. The command:

\begin{verbatim}
docker compose run --rm dev uv run python scripts/verify/paired_analysis.py
docker compose run --rm dev python scripts/paper_figures.py
\end{verbatim}

The first command writes \texttt{arxiv/generated/numbers.json} and
\texttt{arxiv/generated/numbers.tex}, together with the cross-fitted cascade view and
the excluded MiniCheck arm, and it hashes the four prediction files behind those results
into \texttt{arxiv/evidence/raw\_hashes.json}. It reads three further prediction files,
the two superseded pre-parity frontier runs that section 12 uses to bound prompt movement,
one per reasoning effort, and the capped first pass that section 12's same-prompt repeat is
measured against; those three are hashed in \texttt{arxiv/evidence/input\_hashes.json}
instead, because the stamped file cannot be added to without invalidating its proof. The second regenerates the figures from
\texttt{numbers.json} alone.

Both commands are run inside the released package. That package is not hand-assembled: it
is built by \texttt{scripts/build\_public\_export.sh} and published at
\url{https://github.com/adorosario/jev-rag-claim-verification} when this paper is posted.
It holds the raw predictions, the prompts, the pinned model registry, the pricing snapshot
used for every cost figure, the example-ID list, the leaderboard capture, the incident log,
the analysis code and its tests. After assembling the package the builder re-checks a named
path for each of those categories, and compares the prompts directory and the tests
directory against our own file by file, since naming one file out of a directory would let
the rest go missing quietly; it exits non-zero rather than emit a package that fails any of
those checks. A comparison by name is not the same as a suite that runs, so the builder
also runs the shipped tests inside the assembled package and emits nothing if they do not
pass there. Some of those tests read what the package withholds by design, the paper
source, the builder and the vendor's agreement; they ship, so the file-by-file comparison
stays true, and they skip in the package on a marker the builder writes rather than on the
absence of their input. The tests that need the gated dataset skip without a token. The
mapping from the categories listed here to the paths the builder verifies
is held by \texttt{tests/test\_public\_export.py}, so the sentence and the script cannot
drift apart unnoticed. The builder itself is the one script we do not publish, because it
names the internal service that the export exists to strip; what a reader gets instead is
the other half of the check. \texttt{scripts/verify/} \texttt{check\_public\_release.sh} is
in the package, carries the same list of paths, and compares each published file by
\texttt{sha256} against the copy beside it, so the inventory can be confirmed from the
reader's side without trusting ours. It compares bytes rather than reporting an HTTP status
because it once reported a status: a package pushed before a registry correction answered
200 on every path while publishing the superseded value, and a check that asks only whether
a file exists cannot see that. Two other things are held back and neither is evidence for
anything here: an adapter for a private production system that this study does not evaluate,
together with its registry entry and its test, and the internal decision documents. Anything else
listed here that a reader cannot find in that repository is a release error on our part,
and we would rather be told about it than have it excused.

The provenance claim in this paper is one a reader can check without trusting us and
without access to our development history. \texttt{arxiv/evidence/raw\_hashes.json} carries
the sha256 of the four prediction files behind every result in this paper, and
\texttt{arxiv/evidence/raw\_hashes.json.ots} is an OpenTimestamps proof of that file. Both
are in the released package. Running \texttt{ots verify} against them confirms that the
proof commits to this exact file, which is the check that fails the moment the hash file is
edited, and reports the attestations standing behind the commitment. At the time of writing
those are calendar-server attestations pending confirmation in the Bitcoin blockchain;
\texttt{ots upgrade} replaces them with a block attestation once the commitment is mined,
and only then is the time independent of the calendars. Hashing the released prediction
files establishes that they are the files the hash file names, so the pair shows the
predictions were fixed before the analysis that reads them. Our own repository additionally pins the same file with an annotated
git tag, which is bookkeeping for us rather than evidence for a reader: the tag is an
internal reference, the timestamp proof is the artefact that travels. Auxiliary inputs
that are not part of that lock, namely the leaderboard capture, the example-ID list, the
two superseded pre-parity frontier runs and the capped first pass of the low-effort one,
are hashed separately in \texttt{arxiv/evidence/input\_hashes.json}, which carries no
timestamp proof and does not claim one.

\section{Limitations in full}
\label{app:limitations}

Section~\ref{sec:limitations} lists the limitations of this study. This appendix gives each
one its numbers, in the same order. The long-form accounting behind them, including the
family-boundary sensitivity of the correction below, the MiniCheck anchor's full diagnosis
and the same-prompt repeat's per-item reasoning-token coverage, is
\texttt{docs/reference/limitations-in-full.md} in the released package, generated from this
appendix and the same macro file so it cannot disagree with what is printed here.

\paragraph{Indistinguishable is not equivalent.}
The paired difference is ${\JevAstraDeltaSignedPP}$ points on Jev minus Astra, interval
[${\JevAstraDeltaCILo}$, \JevAstraDeltaCIHi] over \BootstrapResamples\ resamples, wide
enough to hold a several-point advantage either way, and the exact McNemar at
$p = \JevAstraMcNemarP$ is a failure to reject and nothing more. We preregistered no
equivalence margin and ran no equivalence test, so no claim of statistical equivalence
appears anywhere here. The thinnest minority class in any component dataset is
\ThinnestMinorityClassN\ claims, which makes the per-dataset cells of
Appendix~\ref{app:per-dataset} descriptive rather than inferential.

\paragraph{The \NClearingUncorrected\ differences that clear both instruments, and what the
correction does to them.}
Section~\ref{sec:setup} fixes the criterion per estimand: on false verification, where the
interval and the test rest on the same gold-unsupported claims, both instruments are
required; on macro-balanced accuracy, where the interval is on the macro average and the
test on pooled per-item correctness, the interval governs and the $p$-value corroborates.
\NClearingUncorrected\ comparisons meet it uncorrected, and \MultiplicitySurvivingN\ of them
survive the Holm-Bonferroni adjustment the criterion also requires.

The false-verification gap between the single systems is \FVRGapPP\ points, interval
[\FVRGapCILo, \FVRGapCIHi] over the \NGoldUnsupported\ gold-unsupported claims, exact
McNemar $p = \FVRGapMcNemarP$ on \FVRJevOnlyFalseVerifies\ and
\FVRAstraOnlyFalseVerifies\ discordant claims, $p = \FVRGapMcNemarPHolm$ adjusted. That one
is real on every reading.

Four cautions attach to the gap between the two single systems. It is a difference in
permissiveness rather than in skill, and any system can lower it by calling fewer claims
supported, which is why Section~\ref{sec:discussion} asks for it to be priced against the
false-alarm rate instead of read alone. It carries the Holm-Bonferroni adjustment
printed above, at \FVRGapMcNemarPHolm\ adjusted against \FVRGapMcNemarP\ raw, which is a
correction over the \MultiplicityFamilyN\ tests this paper reports and not over the wider
space of comparisons these data would support. It is not the same estimand as the accuracy
comparison it is set against: the accuracy tie is a macro average over \NDatasets\ datasets
of roughly \SupportAggreFactCNNN\ claims each, while the gap is a rate pooled over all
\NGoldUnsupported\ gold-unsupported claims with no per-dataset weighting, so pooling gives
it more effective sample and the two are not two readings of one measurement on the same
footing. And it comes from one sample and one pair of model versions, so what the contrast
shows is that these \NPaired\ claims are enough to separate the systems on one quantity and
not on the other, rather than that the two quantities were tested head to head.

The cascade's guard penalty against the high-effort arm is the other survivor,
\CascadeAstraHighFVRDeltaPP\ points, interval [\CascadeAstraHighFVRDeltaCILo,
\CascadeAstraHighFVRDeltaCIHi], median exact McNemar
$p = \CascadeAstraHighFVRMcNemarPMedian$, $p = \CascadeAstraHighFVRMcNemarPHolm$ adjusted.
It is a median rather than a single exact test, and its direction is against the
architecture this paper proposes. The cascade's guard penalty against Astra is
\CascadeAstraFVRDeltaPP\ points, interval [\CascadeAstraFVRDeltaCILo,
\CascadeAstraFVRDeltaCIHi], median $p = \CascadeAstraFVRMcNemarPMedian$ and
$p = \CascadeAstraFVRMcNemarPHolm$ adjusted, so we report it as clearing both instruments
uncorrected and sitting on the $0.05$ boundary once corrected, not as a demonstrated
separation. Dropping the two MiniCheck tests leaves it at
$p = \CascadeGuardHolmPExcludingMiniCheck$ and dropping any single member also leaves it
above $0.05$. Holm multiplies it by the number of tests whose raw $p$ is at least its own,
so the \CascadeGuardHolmMembersAboveN\ members above it in the step-down decide it: it
clears only when the \CascadeGuardHolmBreakDroppedN\ largest of those go, for a family of
\CascadeGuardHolmClearsAtFamilyN, where it adjusts to
$p = \CascadeGuardHolmPAtThatFamily$. Unlike the gap between the single systems, which
holds for any family smaller than \FVRGapHolmBreakFamilyN\ tests, this result turns on
where the family boundary falls, and we report it against the boundary fixed in code before
the adjustment was computed rather than against one that would let it through.

\paragraph{Raw and adjusted $p$ for the whole family.}
Table~\ref{tab:holm} gives every exact McNemar test this paper reports, raw and adjusted.
\MultiplicityMedianMembersN\ of the \MultiplicityFamilyN\ entries are medians over
cross-fitting repeats rather than single exact tests, computed on overlapping resamples of
the same \NPaired\ claims, and a median of $p$-values has no null distribution of its own,
so Holm over a family containing them is descriptive on those members and we claim no error
rate for them. That includes one of the \MultiplicitySurvivingN\ survivors, which is why
Section~\ref{sec:cascade-inference} reports it with the interval beside the test rather than
on the $p$ alone. One further comparison clears one instrument and not the other and is
therefore not counted: the cascade's guard improvement over Jev alone is
${\CascadeJevFVRDeltaPP}$\ points with an interval of [${\CascadeJevFVRDeltaCILo}$,
${\CascadeJevFVRDeltaCIHi}$] that excludes zero, while its median $p$ of
\CascadeJevFVRMcNemarPMedian\ falls below $0.05$ in only
\CascadeJevFVRRepeatsPBelowFive\ of \CrossfitRepeats\ repeats.
\begin{table}[htbp]
\centering\small
\caption{Holm-Bonferroni over the \MultiplicityFamilyN\ exact McNemar tests this paper
reports, the family fixed in code before the adjustment was computed. Rows marked
$\dagger$ are medians over cross-fitting repeats rather than single exact tests. The two
MiniCheck rows are evidence for excluding that arm rather than results about it.}
\label{tab:holm}
\begin{tabular}{lrr}
\toprule
Comparison & raw $p$ & Holm $p$ \\
\midrule
Jev vs.\ Astra, accuracy                      & \JevAstraMcNemarP & \JevAstraMcNemarPHolm \\
Jev vs.\ Astra high effort, accuracy          & \JevAstraHighMcNemarP & \JevAstraHighMcNemarPHolm \\
Astra vs.\ Astra high effort, accuracy        & \AstraAstraHighMcNemarP & \AstraAstraHighMcNemarPHolm \\
Jev vs.\ Astra, false verification            & \FVRGapMcNemarP & \FVRGapMcNemarPHolm \\
Cascade vs.\ Astra, accuracy$^\dagger$        & \CascadeAstraMcNemarPMedian & \CascadeAstraMcNemarPHolm \\
Cascade vs.\ Astra, false verification$^\dagger$ & \CascadeAstraFVRMcNemarPMedian & \CascadeAstraFVRMcNemarPHolm \\
Cascade vs.\ Jev, accuracy$^\dagger$          & \CascadeJevMcNemarPMedian & \CascadeJevMcNemarPHolm \\
Cascade vs.\ Jev, false verification$^\dagger$ & \CascadeJevFVRMcNemarPMedian & \CascadeJevFVRMcNemarPHolm \\
Cascade vs.\ Astra high, accuracy$^\dagger$   & \CascadeAstraHighMcNemarPMedian & \CascadeAstraHighMcNemarPHolm \\
Cascade vs.\ Astra high, false verification$^\dagger$ & \CascadeAstraHighFVRMcNemarPMedian & \CascadeAstraHighFVRMcNemarPHolm \\
Fixed-threshold guard (Table~\ref{tab:random}) & \CascadeAtNinetyNineFVRVsAstraMcNemarP & \CascadeAtNinetyNineFVRVsAstraMcNemarPHolm \\
\midrule
Jev vs.\ MiniCheck & $\JevMiniCheckMcNemarPMantissa \times 10^{-\JevMiniCheckMcNemarPExponent}$ & $\JevMiniCheckMcNemarPHolmMantissa \times 10^{-\JevMiniCheckMcNemarPHolmExponent}$ \\
Astra vs.\ MiniCheck & $\AstraMiniCheckMcNemarPMantissa \times 10^{-\AstraMiniCheckMcNemarPExponent}$ & $\AstraMiniCheckMcNemarPHolmMantissa \times 10^{-\AstraMiniCheckMcNemarPHolmExponent}$ \\
\bottomrule
\end{tabular}
\end{table}

\paragraph{What was selected in sample, and what was not.}
The gating sweep of Section~\ref{sec:calibration}, the tuned cascade of
Section~\ref{sec:tuned} and every row of the confidence column of Table~\ref{tab:random}
are scored on the claims their thresholds were read off. The reportable cascade is the
exception: its threshold is chosen out of fold by \CrossfitRepeats\ repeats of
\CrossfitFolds-fold cross-fitting, so the \SelectionOptimismPP\ points between the
in-sample best (\CascadeInSampleBAccTwoDP) and the cross-fitted estimate is a penalty that
procedure avoids rather than one the headline pays. What cross-fitting does not undo is the
layer outside it: the threshold grid, the decision to route on confidence at all and
macro-balanced accuracy as the objective were fixed with the development outcomes visible.
The selection is unstable at this size, picking one threshold \ThresholdPickedEighty\ times
and a very different one \ThresholdPickedNinetyNine\ times across
\CrossfitThresholdFits\ fold-fits. The gate thresholds in the sweep and the \NDatasets\
per-dataset cells sit outside the Holm family of Section~\ref{sec:setup}, which covers the
exact McNemar tests between the configurations this paper reports, and carry no
multiple-comparison correction. So does the recomputation of the guard penalty against the
superseded pre-parity run: that
recomputation is a sensitivity check on a retracted comparator rather than a result, and is
neither a member of the family nor one of the \NClearingUncorrected\ comparisons that clear
both instruments.

\paragraph{The cascade's sign is unstable in two different ways.}
On the discordant items the exact McNemar works from, the cascade is ahead in
\CascadeAstraMcNemarFavoursCascade\ of \CrossfitRepeats\ repeats and Astra in
\CascadeAstraMcNemarFavoursAstra; the test clears $0.05$ in
\CascadeAstraRepeatsPBelowFive\ of them, all \CascadeAstraRepeatsPBelowFiveForCascade\
favouring the cascade. On the macro-balanced accuracy the paper reports it is not stable:
the cascade lands below Astra in \CascadeAstraRepeatsMacroBelowAstra\ of the repeats and
\CascadeAstraBootstrapDrawsNegativePct\% of the bootstrap draws put the difference below
zero.

\paragraph{One draw of the fold assignment.}
Every cross-fitted mean averages \CrossfitRepeats\ draws of the fold assignment at one
recorded seed, so each carries a Monte Carlo error separate from the sampling uncertainty
over claims: \CrossfitSEMBAccPP\ points on macro-balanced accuracy,
\$\CrossfitSEMCost\ on cost per thousand claims, \CrossfitSEMEscalatedPP\ points on the
escalation share and \CrossfitSEMFVRPP\ points on the false-verification rate. Read those
quantities to that precision rather than to the digits we print. A different seed moves
them together, since they descend from how often the fold-fits choose
\CascadeInSampleThreshold, so this is one draw and not several.

\paragraph{The evaluation sample is a released list, not a rerunnable draw.}
The \NPaired\ example IDs are a fixed list, hashed and released with the predictions, and no
code here regenerates the draw, so a reader can check exactly what was evaluated and cannot
re-derive how it was chosen. The list came out of an exploratory run of the decision model in
which every retained row succeeded, so this sample cannot speak to that system's failure
rate, and the absence of model-caused failures in Section~\ref{sec:setup} is an independent
observation only for the frontier configurations and MiniCheck. No claim here rests on
either system's failure rate.

\paragraph{The cascade is an offline replay.}
Both systems answered all \NPaired\ claims independently and the cascade is composed from
those stored predictions. No claim was re-sent and no deployed router was built. A live
implementation adds a second network hop on the escalated fraction, with queueing and
failure behaviour this study does not measure.

\paragraph{Every cost figure is two price lists on one day.}
Costs are measured token counts at list prices from two vendors' public pages on a single
date, archived and hashed as \texttt{configs/} \texttt{pricing\_2026-09-20.yaml} in the
released package. Nothing was invoiced, so these are modelled costs. Three consequences:
prices move, and a ratio from a snapshot ages the moment either vendor reprices, which is
why the dated file ships rather than the ratio being asserted; list price is not what a
buyer at volume pays, and negotiated rates are public for neither vendor and would not move
the arms by the same factor; and the decision model's free output tokens are its vendor's
commercial policy, not a property of the architecture, so reading
\CostRatioAstraOverJev$\times$ as durable would be reading a price sheet as an engineering
result. What survives repricing is that a binary decision need not be billed at generation
rates.

\paragraph{Each arm was run once.}
Every number comes from a single pass of each configuration at the provider's default
sampling settings, and no repeatability study was run. One same-prompt repeat of the
frontier arm exists in the released artifacts: the discarded first pass of
Appendix~\ref{app:incidents} and the pass that replaced it carry the same prompt hash,
reasoning effort and model, differing only in our completion-token cap, raised from
\SamePromptRepeatCappedTokenCap\ to \SamePromptRepeatTokenCap. That shared prompt is the
superseded pre-parity wording rather than the corrected prompt behind every frontier number
here, so the bound is measured on a prompt no reported configuration used. On the
\SamePromptRepeatCompared\ claims the capped pass answered, the two agree on every verdict:
\SamePromptRepeatFlips\ flips and macro-balanced accuracy \SamePromptRepeatBAcc\ against
\SamePromptRepeatBAccOther, a difference of \SamePromptRepeatDeltaPP. It is a narrow bound. Those rows are the ones whose answers fitted
the smaller budget, so they are the short-answer end of the sample and the end where the model
reports zero reasoning tokens \SamePromptRepeatComparedReasoningZero\ times; the remaining
\SamePromptRepeatExcluded\ claims are untested for repeatability, and they are where every
prompt-parity flip landed. No comparable repeat exists for the high-effort configuration.

For the decision model an earlier pass over the identical example IDs is on disk and is the
file the example-ID list is drawn from. Between it and the run behind this paper,
\JevRerunFlips\ of \NPaired\ verdicts changed and macro-balanced accuracy rose from
\JevRerunBAccTwoDP\ to \JevBAccTwoDP, \JevRerunDeltaPP\ points. Both passes sit below the
frontier arm's \AstraBAccTwoDP, so the movement widens the deficit rather than reversing
anything; what it does is exceed the headline gap of \JevAstraDeltaPP\ points that
Section~\ref{sec:same-score} reports.

On the frontier arm we varied the prompt once, in the parity correction of
Appendix~\ref{app:incidents}, and every run on both sides of that edit is released. Two runs
of a sampled model differ by the edit and by a fresh draw at once, and this study cannot
separate them: the repeat above covers only rows where the model does no sampled reasoning,
and all \PromptParityFlipsInCappedNoAnswer\ of the \PromptParityFlips\ flips the edit moved
fall outside those rows. The direction is worth printing because it is not the direction the
edit pushes: the added trigger makes \textsc{not\_supported} easier to reach, and
\PromptParityFlipsToUnsupported\ of the flips went that way against
\PromptParityFlipsToSupported\ the other. On the low-effort arm the edit moved
\PromptParityFlips\ of \NPaired\ verdicts and macro-balanced accuracy from
\PromptParityPreBAcc\ to \PromptParityPostBAcc, all \PromptParityFlipsOnSupported\ of them on
gold-supported claims and \PromptParityFlipsOnUnsupported\ on gold-unsupported ones, so it
moved the false-verification gap by \PromptParityFVRGapShiftPP\ points and that arm's rate
stands at \PromptParityPreFVR\% before the change and \PromptParityPostFVR\% after.

That is the low-effort arm and it does not speak for the other. The same edit moved
\PromptParityHighFlips\ verdicts on the high-effort arm,
\PromptParityHighFlipsOnUnsupported\ of them on a gold-unsupported claim, taking that arm's
false-verification rate from \PromptParityHighPreFVR\% to \PromptParityHighPostFVR\% and its
macro-balanced accuracy from \PromptParityHighPreBAcc\ to \PromptParityHighPostBAcc. That
rate is one leg of the guard penalty that is one of the \MultiplicitySurvivingN\ survivors,
so rather than certify the survivor as unaffected we recomputed it against the run the
correction replaced: \CascadeAstraHighFVRPreParityDeltaPP\ points there against
\CascadeAstraHighFVRDeltaPP\ reported, median exact McNemar
$p = \CascadeAstraHighFVRPreParityMcNemarPMedian$, and all
\CascadeAstraHighFVRPreParityRepeatsPBelowFive\ repeats separating on either run. The
correction moved that result towards the smaller number, which is the one we report. The
decision model's prompt was never varied at all.

What the movement did disturb is the sign of the headline comparison. Jev scores
\JevBAccTwoDP. Against the pre-correction frontier run at \PromptParityPreBAcc\ the decision
model led; against the corrected run at \PromptParityPostBAcc\ the ordering reverses, to
${\JevAstraDeltaSignedTwoDP}$ points on Jev minus Astra. So a correction to our own prompt,
made for a reason unrelated to the outcome, moved the comparator further than the entire
difference this paper reports and reversed its direction. That is the strongest available
illustration of this paper's own thesis, and both runs are released so a reader can check
the ordering.

\paragraph{Latency was not measured under control.}
The bulk runs were issued concurrently, so their per-request timings describe our harness as
much as either system. No latency figure or ratio appears in this paper, and the controlled
single-request protocol the project specifies was not run.

\paragraph{Contamination cannot be ruled out.}
LLM-AggreFact and its components have been public long enough that any system here may have
seen them in training, and we cannot inspect either vendor's training data. A
contamination-resistant holdout of newly written claims exists for this project and is not
part of this study.

\paragraph{One frontier family, two reasoning efforts.}
Both efforts ran under one prompt, byte for byte, which ends by forbidding reasoning in the
response (Appendix~\ref{app:prompts}), though not under one completion-token cap,
\AstraTokenCap\ against \AstraHighTokenCap, because the cap covers the reasoning too and the
low one would have truncated the high arm. On the convention of
Table~\ref{tab:paired-contrasts} the two efforts differ by
${\AstraAstraHighDeltaSignedPP}$ points at exact McNemar $p = \AstraAstraHighMcNemarP$, while
the reasoning the provider reports moves from a median of \AstraReasoningMedian\ tokens to
\AstraHighReasoningMedian. So the contrast bounds what the effort parameter buys, with the
headroom it needs, under that output-format instruction, and it is not a test of a prompt
that invites reasoning: the provider-default configuration was never run. A single family at
a single point in time supports no statement about frontier models in general.

\paragraph{An arm we ran and excluded.}
We ported MiniCheck \citep{tang2024minicheck} from the authors' public release
\citep{tang2024aggrefactrepo} as a third system and it scored \MiniCheckBAcc\
[\MiniCheckBAccCILo, \MiniCheckBAccCIHi] macro-balanced accuracy on the same \NPaired\
claims. That is not a threshold artifact: sweeping \MiniCheckCutpointsSwept\ cutpoints buys
\MiniCheckRetuningHeadroomPP\ points at the best reachable cut, still
\MiniCheckGapToJevAtBestCutPP\ points below Jev. It was also the external anchor for harness
validity, and the published LLM-AggreFact figure is \MiniCheckPublishedTestBAcc\ on the test
split against our \MiniCheckBAcc\ on the dev claims, a shortfall of
\MiniCheckShortfallVsPublishedPP\ points that our own interval's half-width of
\MiniCheckBAccCIHalfWidthPP\ points cannot absorb. The conclusion runs both ways: we exclude
the arm, and we do not claim the rest of the harness is sound. What the failure implicates is
narrower than the whole harness. Not the metric layer, a second implementation cross-checked
against established libraries, which produces the Jev and Astra rows; not the gold labels or
the join, since all four arms are scored against one gold vector built once from the pinned
parquet, so an error there would move every arm together. What it does implicate is
everything specific to our adapter: chunking, evidence truncation and the aggregation of
per-chunk entailment scores, none of which the API arms use.

\paragraph{The published reference we quote is a partial capture.}
The file we take that reference from holds \PublishedLeaderboardRows\ rows, and quoting only
the one that excuses our port would be the selective reading Section~\ref{sec:tuned} objects
to, so here are all of them on the test split: Bespoke-MiniCheck-7B
\PublishedBespokeMiniCheckTestBAcc, Claude-3.5-Sonnet
\PublishedClaudeThreeFiveSonnetTestBAcc, \texttt{gpt-4o-2024-05-13}
\PublishedGptFourOTestBAcc, MiniCheck-Flan-T5-L \PublishedMiniCheckFlanTFiveLTestBAcc. It is
a capture made to pin the harness anchor, not a survey: the leaderboard lists systems we did
not keep, and the rows we kept were not chosen by any stated rule, so these figures place
nobody in a field and bound nothing. Nor are they a comparison with our results, which are
macros over \NPaired\ dev claims under one prompt written to be semantically identical across
arms. They do not establish that our short-answer prompt cost the frontier arm points, and
they do not establish that it did not.

\paragraph{The calibration comparison has one participant.}
Jev returns a native probability. The provider of GPT-6 Astra rejects the logprobs parameter
for this model, recorded verbatim in the released \texttt{runs/model\_live\_check.json}, and
no self-reported confidence was elicited, so the frontier configurations have no probability
of any provenance. Every calibration statement here is about Jev, and the asymmetry is
architectural rather than a gap in our measurement. Section~\ref{sec:calibration} reports
Jev's negative log likelihood as \JevNLL\ and says most of it is the clip:
\JevNLLClipN\ claims carry a reported probability at the boundary on the wrong side of the
gold label, each charged \JevNLLClipNatsEach\ nats at the constant
\texttt{configs/benchmark.yaml} marks provisional. Table~\ref{tab:logloss-eps} re-scores the
identical predictions at other constants. A constant one-half forecast scores
\JevNLLConstantHalf, so the reported figure exceeds that reference only for constants below
$\JevNLLCrossoverEpsMantissa \times 10^{-\JevNLLCrossoverEpsExponent}$, which is finer than
anything this vendor reports, $p_{\mathrm{supported}}$ arriving at
\JevProbDecimals\ decimals with \JevProbAtZeroN\ literal zeros and \JevProbAtOneN\ literal
ones, which is why Section~\ref{sec:calibration} reads the verdict off the bounded scores
instead.

\begin{table}[t]
\centering\small
\caption{Jev's negative log likelihood over clipping constants, same predictions and same
labels each time. The first row is the configured value and the figure
Section~\ref{sec:calibration} reports; the last is half the step the vendor reports the
probability on, which is the coarsest constant that still charges a reported zero more
than a reported \JevProbStep. The reference a constant one-half forecast scores is
\JevNLLConstantHalf.}
\label{tab:logloss-eps}
\begin{tabular}{lr}
\toprule
Clipping constant & Jev NLL \\
\midrule
$\JevNLLClipEpsMantissa \times 10^{-\JevNLLClipEpsExponent}$ (configured) & \JevNLL \\
$\JevEpsGridMicroMantissa \times 10^{-\JevEpsGridMicroExponent}$ & \JevNLLAtEpsMicro \\
$\JevEpsGridMilliMantissa \times 10^{-\JevEpsGridMilliExponent}$ & \JevNLLAtEpsMilli \\
\JevProbHalfStep\ (half the reported step) & \JevNLLAtHalfStepClip \\
\bottomrule
\end{tabular}
\end{table}

\paragraph{No example was excluded for length, and nothing was truncated.}
The longest input is \JevMaxInputTokens\ tokens for the decision model and
\AstraMaxInputTokens\ for the frontier arm, well inside both documented context limits, and
all \NPaired\ rows of all four prediction files carry a successful status. Nothing was dropped
for exceeding a context window and no evidence text was shortened to fit one. That also bounds
what the sample can say: a set of claims that never approaches either limit cannot tell a
reader how either system behaves on the long documents that would.

\paragraph{The harness was built with AI coding assistance.}
Claude Code wrote most of the analysis and adapter code under human direction. What stands
behind the numbers is this. The metric layer is a second implementation:
\texttt{src/metrics.py} imports no scikit-learn and computes every metric directly, and
\texttt{tests/test\_metrics\_crosscheck.py} recomputes each with scikit-learn on randomised
fixtures and requires agreement to \texttt{1e-12}, negative log loss excepted at
\texttt{1e-9} because the two clip differently, so a bug would have to exist identically in
two implementations to survive. Expected calibration error has no library equivalent and is
covered by unit tests on constructed cases. The array shortcuts inside the bootstrap and
cross-fitting loops are asserted equal to that metric code before any result is reported.
Each prediction file was committed when its run finished and has not changed since: the
analysis re-derives every input's sha256 on every run and aborts rather than rewrite the
stamped evidence lock, so an input edited after the lock fails the run instead of quietly
moving a number. What we do not claim is a second implementation of the pipeline above the
metrics: one script turns the locked predictions into every number here.

\paragraph{The TypeSafe record, in order.}
Our protocol cited a clause restricting publication of benchmark results. It is not in the
live agreement. The citation traces to a deep-research summary in this project's research
directory, which the project's own README marks as not a source of truth. The founder decision
dated 2026-09-18, in the released package, accepted a breach risk on that premise and now
carries a correction header, because the premise did not hold. On 2026-09-20 we fetched the
live agreement, captured it, hashed the capture and searched it: the words
\texttt{benchmark} and \texttt{publish} do not occur, and the subsection our protocol cited,
2.3(f), is about interfering with the operation of the service. The decision that followed is
Alden's of 2026-09-21, recorded in the package's permission README: publish, and request no
permission. The request drafted under the old premise was never sent and ships headed with
the reason it must not be sent as written. The hash covers our own capture of a page that had
already changed version line between the protocol's reading and ours, so it fixes the bytes
our excerpt came from and nothing else; a reader who wants to check the finding fetches the
clause rather than matching a digest. Neither document is a statement by the vendor.